\documentclass[lettersize,journal]{IEEEtran}

\usepackage{amsmath, amsthm,amssymb,subfigure,cite,color,booktabs,topcapt,times,multirow}
\usepackage[T1]{fontenc}
\usepackage{enumerate}
\usepackage{cite}
\usepackage{url}
\usepackage{algpseudocode}
\usepackage{algorithmicx}
\usepackage{gensymb}
\usepackage[linesnumbered,ruled,vlined]{algorithm2e}
\usepackage{amssymb}
\usepackage{courier}
\usepackage{soul}
\usepackage{amsfonts,amssymb}
\usepackage{lineno,amsmath}

\usepackage{booktabs}
\usepackage{multirow}
\usepackage{pifont}
\usepackage{makecell}

\usepackage{array}
\usepackage[caption=false,font=normalsize,labelfont=sf,textfont=sf]{subfig}
\usepackage{textcomp}
\usepackage{stfloats}
\usepackage{verbatim}
\usepackage{graphicx}
\begin{document}

\title{Bi-AM-RRT*: A Fast and Efficient Sampling-Based Motion Planning Algorithm in Dynamic Environments}

\author{\IEEEauthorblockN{Ying~Zhang,
						Heyong~Wang,
						Maoliang~Yin,
                        Jiankun Wang,~\IEEEmembership{Senior Member,~IEEE,} \\
                        and Changchun Hua,~\IEEEmembership{Senior Member,~IEEE}
                        }

\thanks{This work was supported in part by the National Natural Science Foundation of China under Grant No. 62203378, 62203377, U22A2050, in part by the Hebei Natural Science Foundation under Grant No. F2022203098, F2021203054, in part by the Science and Technology Research Plan for Colleges and Universities of Hebei Province under Grant No. QN2022077, and in part by the Hebei Innovation Capability Improvement Plan Project under Grant No. 22567619H. \emph{(Corresponding author: Ying Zhang.)}}

\thanks{Y. Zhang, H. Wang, M. Yin, and C. Hua are with the School of Electrical Engineering and the Key Laboratory of Intelligent Rehabilitation and Neromodulation of Hebei Province, Yanshan University, Qinhuangdao, 066004, China. (e-mail: yzhang@ysu.edu.cn; wtk0405@163.com; yin924431601@163.com; cch@ysu.edu.cn).}

\thanks{J. Wang is with the Shenzhen Key Laboratory of Robotics Perception and Intelligence, Shenzhen 518055 China, and also with the Department of Electronic and Electrical Engineering, Southern University of Science and Technology, Shenzhen 518055, China (e-mail: wangjk@sustech.edu.cn).}
}

\markboth{IEEE Transactions on Intelligent Vehicles}%
{Shell \MakeLowercase{\textit{et al.}}: Bi-AM-RRT*: A Fast and Efficient Sampling-Based Motion Planning Algorithm in Dynamic Environments}

\maketitle

\begin{abstract}
The efficiency of sampling-based motion planning brings wide application in autonomous mobile robots. 
The conventional rapidly exploring random tree (RRT) algorithm and its variants have gained significant successes, 
but there are still challenges for the optimal motion planning of mobile robots in dynamic environments.
In this paper, based on Bidirectional RRT and the use of an assisting metric (AM), we propose a novel motion planning algorithm, namely Bi-AM-RRT*. 
Different from the existing RRT-based methods, the AM is introduced in this paper to optimize the performance of robot motion planning in dynamic environments with obstacles.
On this basis, the bidirectional search sampling strategy is employed to reduce the search time. 
Further, we present a new rewiring method to shorten path lengths. The effectiveness and efficiency of the proposed Bi-AM-RRT* are proved through comparative experiments in different environments. Experimental results show that the Bi-AM-RRT* algorithm can achieve better performance in terms of path length and search time, and always finds near-optimal paths with the shortest search time when the diffusion metric is used as the AM.
\end{abstract}


\section{Introduction}
Recent advances in robotics have prompted an increasing number of autonomous mobile robots to be used in various fields, such as transportation \cite{chen2023milestones}, manufacturing \cite{zhao2022multimobile}, rescue \cite{zhu2021novel}, domestic service \cite{zhang2023semantic}, and so on.
As a fundamental task of mobile robots, motion planning aims to plan a feasible collision-free path from the starting point to the goal point for the robot in the working environment with static or dynamic obstacles \cite{zhang2022receding}. In such context, lots of research efforts have been conducted on the motion planning problem.
For instance, based on the grid map, the Dijkstra \cite{dijkstra1959note} algorithm can derive a feasible trajectory by traversing the entire map. In order to save computing resources, A* \cite{hart1968formal} and anytime repairing A* (ARA*)\cite{likhachev2003ara} use a heuristic search strategy to quickly obtain optimal solution. 
However, these methods are not suitable for high-dimensional environments or differential constraints.
Moreover, to address dynamic obstacles, the D* \cite{stentz1995focussed} and the anytime D* \cite{likhachev2005anytime} are investigated to search for feasible solutions in dynamic environments. The methods above are grid-based algorithms that require discretization of the state space, which leads to an exponential growth in time spent and memory requirements 
with the increase of the state space dimension \cite{wang2022gmr}. To reduce the time cost and memory usage, 
diffusion map is employed \cite{coifman2006diffusion}. It is a non-linear dimensionality reduction technique, and seeks for a feasible solution by transforming each state on the map into a diffusion coordinate \cite{chen2016motion}.
Nevertheless, this treatment tends to ignore some details in the environment, leading to poor planning performance or even getting into trap in complex dynamic environments. %

For fast and high-quality motion planning in complex dynamic environments, sampling-based methods have attracted significant attention. Typically,
the rapidly exploring random tree (RRT) algorithm \cite{lavalle1998rapidly} has been widely used and achieved great success due to its efficiency and low memory usage. To this end, many of its variants have been presented.
For example, RRT-connect \cite{kuffner2000rrt} shortens the search time by exploiting goal bias and using two trees to search simultaneously. RRT* \cite{karaman2011anytime} adds a rewiring process to shorten the path length. 
Extended-RRT \cite{li2012extended} re-searches for new collision-free path from the root when there are obstacles in the planned trajectory. But this practice is time-spending.
Besides, the RT-RRT*\cite{naderi2015rt} retains information about the whole tree from the robot's current position, and uses existing branches around obstacles to locally plan feasible paths. However, the growth of the whole tree takes more time.
In such case, an extended RRT-based planning method with the assisting metric (AM) \cite{armstrong2021rrt} is investigated to guide the growth of the tree to shorten the path search time.
Although the utilization of AM can accelerate the RRT exploration process, the search time and path length needs to be improved in dynamic environments. 

In this paper, we propose a novel motion planning method based on bidirectional RRT and AM, namely Bi-AM-RRT*, to reduce the search time and path length in dynamic environments.
The presented Bi-AM-RRT* exploits the trunk information of the reverse tree with the forward tree to efficiently
generate a feasible path to the goal position. 
Based on this, the AM is used to improve the performance of motion planning in environments with obstacles.
The AM can be any metric, such as Euclidean metric, diffusion metric, or geodesic metric.
Besides, in order to optimize the search path, a new rewiring strategy based on the root and goal is presented to shorten the path length. 
The main contributions of this work include:
\begin{itemize}
  \item an AM-based bidirectional search sampling framework for robot motion planning in dynamic environments;
  \item a novelly fast and efficient motion planning algorithm, namely Bi-AM-RRT*, to improve the motion planning performance;
  \item a new rewiring strategy to accelerate the path optimization process to reduce the path length;
  \item evaluation and discussion on comparative experiments in different environments, which demonstrate the validity and efficiency of Bi-AM-RRT*.
\end{itemize}

The remainder of this paper is structured as follows. Section \ref{section2} presents the related work. In Section \ref{section3}, the problem definition and AM-RRT* are introduced. Section \ref{section4} elaborates the proposed Bi-AM-RRT*. Section \ref{section5} and Section \ref{section6} describe the extensive experiments and discuss the results, respectively. Section \ref{section7} concludes this paper.

\section{Related work} \label{section2}
Robot motion planning aims at planning a feasible path for robots, and has received significant attention over the years, especially in dynamic environments. Many algorithms have been proposed to address the motion planning problem.

To plan a feasible path, the artificial potential field algorithm was introduced for robot motion planning \cite{khatib1985real}, which uses the direction of the fastest potential field decline as the moving direction of the robot. However, when in an environment with obstacles, such solution is prone to fall into local optimisation. In recent years, the learning-based motion planning strategies have been investigated.
Everett \emph{et al}. \cite{everett2018motion} proposed an obstacle avoidance method that trains in simulation with deep reinforcement learning (RL) without requiring any knowledge of other agents' dynamics. 
Similarly, Wang \emph{et al}. \cite{wang2020mobile} designed an RL-based local planner, which adopts the global path as the guide path to adapt to the dynamic environment. To optimize the planner, P{\'e}rez-Higueras \emph{et al}. \cite{perez2018teaching} combined inverse reinforcement learning with RRT* to learn the cost function. The introduction of a machine learning improves the agent's path planning and obstacle avoidance performance in dynamic environments.
Notably, these learning-based methods need to train the model in advance, which is time-spending. Moreover, in order to find the optimal trajectory, grid-based motion planning research efforts have been conducted extensively. For example,  based on the grid map, A*\cite{hart1968formal} was used to search for feasible solutions and gained great success. Koenig \emph{et al}. presented D*-lite for robot planning in unknown terrain based on the lifelong planning A*. The performance is closely related to the degree of discretization of the state space. Although these grid-based approaches can always search for the optimal path (if one exists), they do not perform well as the scale of the problem increases, such as time-consuming and high memory consumption.

To improve planning performance, sampling-based methods are considered as a promising solution. In particular, RRT-based algorithms are widely popular due to their ability to efficiently search state spaces and have proven to be an effective way to plan a feasible path for robots \cite{wang2021survey}.
For instance, Kuwata \emph{et al}. \cite{kuwata2008motion} proposed CL-RRT for motion planning in complex environments. This method uses the input space of a closed-loop controller to select samples and combines effective techniques to reduce the computational complexity of constructing a state space. Based on the probabilist collision risk function, Fulgenzi \emph{et al}.
\cite{fulgenzi2010risk} introduced a Risk-RRT method. In this solution, a Gaussian prediction algorithm is used to actively predict the moving obstacles and the sampled trajectories to avoid collisions. To achieve dynamic obstacle avoidance, Naderi \emph{et al}. \cite{naderi2015rt} designed a RT-RRT* algorithm that interweaves path planning with tree growth and avoids waiting for the tree to be fully constructed by moving the tree root with the agent. Analogously, Armstrong \emph{et al}. \cite{armstrong2021rrt} put forward an AM-RRT* by using AM to accelerate the path planning process of RT-RRT*. However, the planning performance is not satisfactory, especially in terms of search time.
In order to reduce the search time, bidirectional search strategies are widely employed. As an early proposed bidirectional tree algorithm, RRT-connect \cite{kuffner2000rrt} uses a greedy heuristic to guide the growth of two trees, thereby shortening the search time. 
Subsequently, other variants such as Informed RRT*-connect \cite{mashayekhi2020informed}, B2U-RRT \cite{wang2021efficient}, Bi-Risk-RRT \cite{ma2022bi}, etc. were proposed. They all incorporate bidirectional search strategy and demonstrate the effectiveness of it.
Inspired by AM-RRT* and RRT-connect, a novel AM-based bidirectional search sampling framework for motion planning, i.e., Bi-AM-RRT*, is proposed in this paper to further reduces the search time and the path length.

Additionally, it is necessary to obtain an optimal path while maintaining the speed of the planner to guarantee the quality of planning. To address the problem of path optimization, Karaman \emph{et al}. \cite{karaman2010incremental} proposed RRT* by using newly generated nodes to rewire adjacent vertices to ensure asymptotic optimality. But the convergence speed is slow. In order to accelerate the convergence speed, Yi \emph{et al}. \cite{yi2018generalizing} suggested a sampling-based planning method with Markov Chain Monte Carlo for asymptotically-optimal motion planning. Chen \emph{et al}. \cite{chen2018fast} designed DT-RRT to abandon the rewire process and add re-search parent based on the shortcut principle. Although this approach can speed up the convergence process, it tends to produce suboptimal paths. Analogously, Wang \emph{et al}. \cite{xinyu2019bidirectional} presented a P-RRT*-connect algorithm to accelerate the convergence of RRT* using an artificial potential field method. Besides, based on the path optimization and intelligent sampling techniques, Islam \emph{et al}. \cite{islam2012rrt} proposed RRT*-Smart, which aims to obtain an optimum or near optimum solution.
Gammell \emph{et al}. \cite{gammell2014informed} investigated the optimal sampling-based path planning with a focused sampling method and presented Informed RRT* to improve the covergence of RRT*. However, these methods face challenges for the efficiency of motion planning in dynamic environments. 

In this paper, based on bidirectional search sampling strategy, a novel motion planning method, namely Bi-AM-RRT*, is proposed with a new rewiring scheme to reduce the path length and the search time for agent to find the goal in dynamic environments.

\section{Preliminaries} \label{section3}
In this section, the problem definition of motion planning and the programs on which the algorithm depends are introduced first, and then the sampling-based planning algorithm with AM, referred to as AM-RRT*, is described.

\subsection{Motion Planning Problem Definition}
Let us define the state space as \textit{X $\in\mathbb{R}^{d}$}. \textit{X$_{obs}\in$ X} denotes obstacles in the state space, while \textit{X$_{free} = X/X_{obs}$} is the free space without obstacles. \textit{x$_{agent} \in X_{free}$} is defined as the state of the mobile robot in the space, and the goal state is represented as \textit{x$_{goal} \in X_{free}$}. In this paper, a search tree \textit{T $\in X_{free}$} is used to generate a feasible collision-free path (i.e., points on the path \textit{x$_{i} \in$ T}) from the start point \textit{x$_{root}$} to the goal point \textit{x$_{goal}$}.
During exploration, the Bi-AM-RRT* can grow both forward tree and reverse tree, which are denoted \textit{T$_{f}$} and \textit{T$_{r}$}, respectively.
In addition, there are user-defined the maximum edge length \textit{e$_{max}$} and the maximum number of nodes \textit{n$_{max}$} in the circular domain with radius \textit{e$_{max}$} to control the growth state of \textit{T}. Let \textit{t$_{exp}$} be the tree growth time. Meanwhile, the root rewiring time and goal rewiring time are denoted as \textit{t$_{root}$} and \textit{t$_{goal}$}, respectively.
When the Euclidean distance is less than $\sigma$ and there is no obstacle between forward tree and reverse tree, two trees can be joined as one tree, where $\sigma$ represents the connecting distance of two trees.

In the presented method, AM \textit{d$_{A}$} can be the Euclidean metric, the diffusion metric \cite{chen2016motion}, and the geodesic metric \cite{owen2010fast}, which are indicated as \textit{d$_{E}$} , \textit{d$_{D}$} , and \textit{d$_{G}$}, respectively. These metrics are used to calculate the distance information of two states in the state space. Specifically, the Euclidean distance is expressed as
\begin{equation}\label{1}
	d_E(x_a, x_b) = \|x_a, x_b\|.
\end{equation}
The Euclidean distance between the two points is obtained based on the L2 norm of \textit{x$_{a}$} and \textit{x$_{b}$}. The diffusion distance is yielded by calculating the Euclidean distance of the approximate diffusion coordinates corresponding to each of the two states, and is described as
\begin{equation}\label{2}
	d_D(x_a, x_b) = \|h(g(x_a)), h(g(x_b))\|
\end{equation}
where \textit{g($\cdot$)} refers to mapping a state $\cdot$ in the grid to the nearest point, and \textit{h($\cdot$)} is mapping a state $\cdot$ to the diffuse coordinates. \textit{d$_{D}$} can provide a good approximation when an obstacle is present. The geodesic distance is the use of the Dijkstra\cite{dijkstra1959note} method to generate a distance matrix from the connection matrix of discretization state space. It has the advantage of high precision, but is time-spending.

Next, the procedures on which the algorithm depends are described \cite{armstrong2021rrt}.
\textit{Cost(T, x)} refers to the length of the path from the root to \textit{x} in \textit{T} based on Euclidean distance. \textit{Path(T, x)} refers to returning the sequence of path nodes from the root to \textit{x} in \textit{T}. \textit{FreePath(x$_{a}$, x$_{b}$)} returns true when there are no obstacles between x$_{a}$ and x$_{b}$. \textit{Nearest(T, x)} returns the nearest neighbor to \textit{x} if there is no obstacle between \textit{x} and its nearest neighbor. \textit{RewireEllipse(T, x$_{goal}$)} is to return the state set within the rewire ellipse \cite{gammell2014informed}. \textit{Enqueue(Q, x)} is the addition of \textit{x} to the end of \textit{Q}. 
\textit{Dequeue(Q)} refers to removing and returning the first item in \textit{Q}. \textit{Push(S, x)} is to add \textit{x} to the front of \textit{S}. \textit{Nearby(T,x)} returns the set of all nodes within E-distance e$_{max}$ of \textit{x}. 
\textit{Pop(S)} refers to deleting and returning the first item in \textit{S}. \textit{Second(S)} means that the second item in \textit{S} is returned but not removed. \textit{UpdateEdge(T, x$_{new}$, x$_{child}$)} replaces the edge \textit{(x$_{parent}$, x$_{child}$)} with \textit{(x$_{new}$, x$_{child}$)} in \textit{T}, where \textit{x$_{parent}$} is the parent of \textit{x$_{child}$} in \textit{T}. \textit{Len(X)} returns the queue length of \textit{X}.  

\subsection{AM-RRT*}
\begin{figure}
	\centering
		\subfigure[]{
		\includegraphics[width=1.6in]{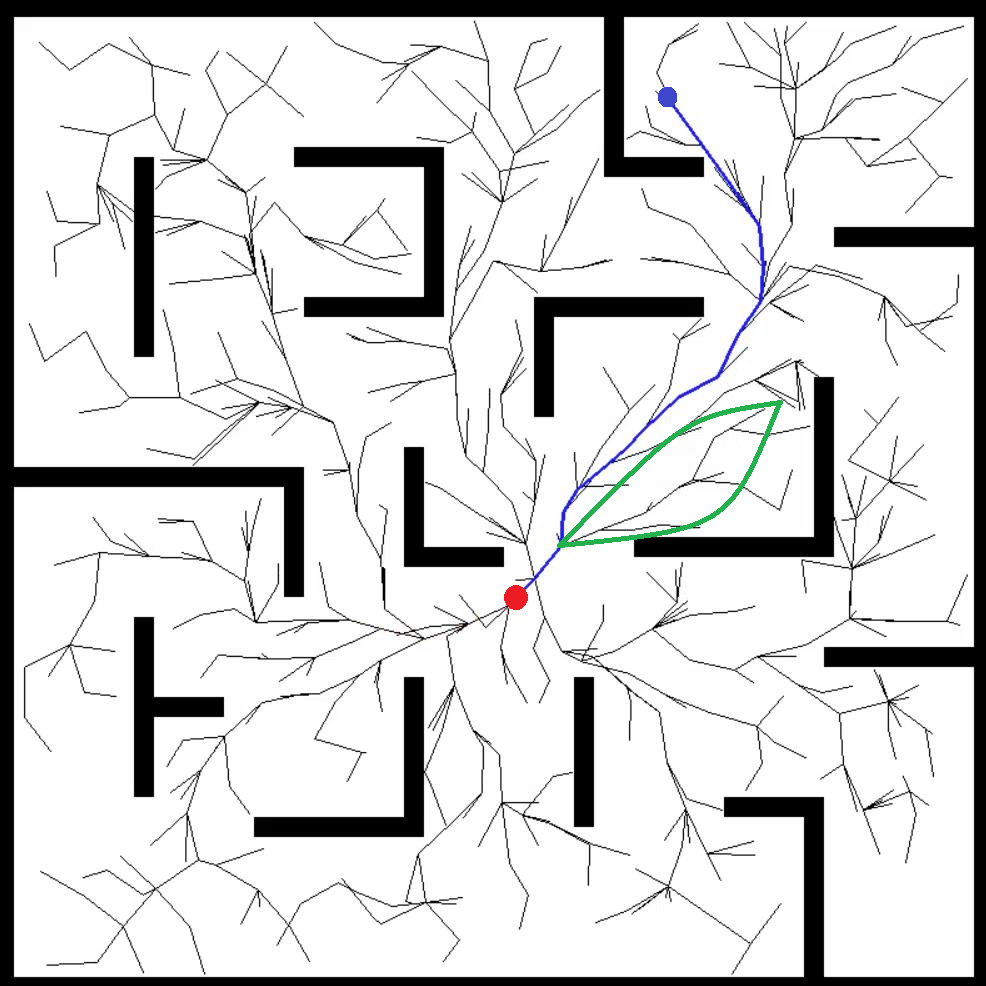}
		}
		\subfigure[]{
		\includegraphics[width=1.6in]{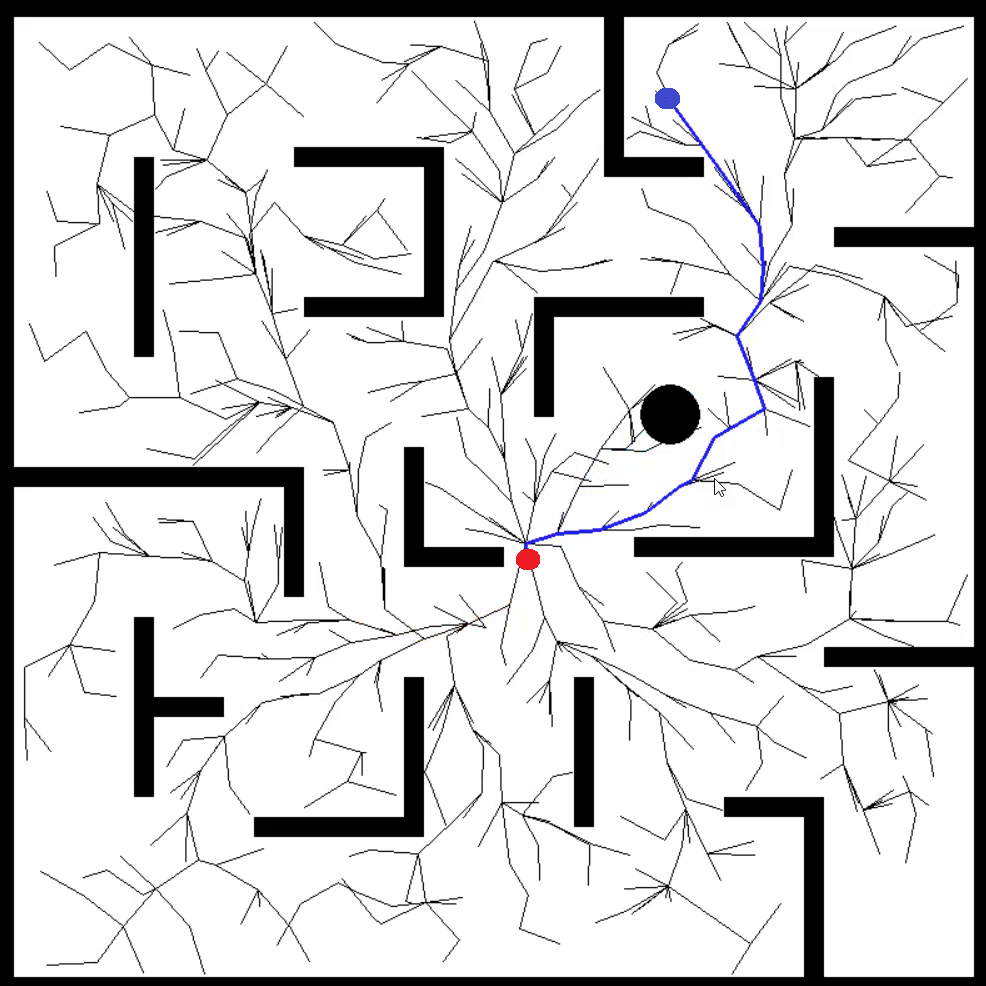}
		}	
	\caption{The obstacle avoidance process of AM-RRT*. (a) The branch information (green circular area) is used for motion planning, and (b) obstacle avoidance when encountering a dynamic obstacle, where red represents the agent, blue represents the goal point, and black represents the obstacle. 
}\label{fig1}
\end{figure}

AM-RRT* is an informed sampling-based planning algorithm with AM \cite{armstrong2021rrt}. Typically, 
AM-RRT* uses the diffusion distance as an AM, which is derived from the diffusion map \cite{coifman2006diffusion} and is also a kind of grid map. It utilizes a dimensional collapse method to reduce time and memory consumption. Although the diffusion distance alone performs poorly in complex environments, but it can achieve good performance as an AM of RRT*,
and can quickly find collision-free paths when obstacles appear.
Fig. \ref{fig1} shows the obstacle avoidance performance with AM-RRT*. When the obstacle appears on the path, AM-RRT* does not regenerate the tree, but uses the information in the whole tree for obstacle avoidance action, especially the node information around the obstacle.
As can be viewed in Fig. \ref{fig1}, when obstacles appear, with the help of diffusion metric, a feasible path can be quickly drawn based on node rewiring by using the branch information of the tree below the original path. During agent movement, the tree is maintained in real time. At the same time, the planned path is also rewired for optimization, and the path lengths are made approximately optimal by successive iterations. The whole process of tree growth is similar to RRT* and its variants. In this process, the diffusion map only plays a leading role in guiding the tree to find the goal point quickly and cover the full space faster while maintaining its probabilistic completeness in a complex environment \cite{armstrong2021rrt}.

\section{Proposed Bi-AM-RRT*}\label{section4}
This paper proposes the Bi-AM-RRT* for real-time optimal motion planning of mobile robots in dynamic environments. 
Generally, our proposed Bi-AM-RRT* uses bidirectional trees (i.e., forward and reverse trees) for searching and accelerates the path optimization by a new rewiring process.
In this section, the details of the of our Bi-AM-RRT*, especially the proposed bidirectional search strategy and the new path rewiring strategy, are presented.

\subsection{Bidirectional Search Strategy for Bi-AM-RRT*}
\begin{figure}
	\centering
	\subfigure[]{
		\includegraphics[width=1.6in]{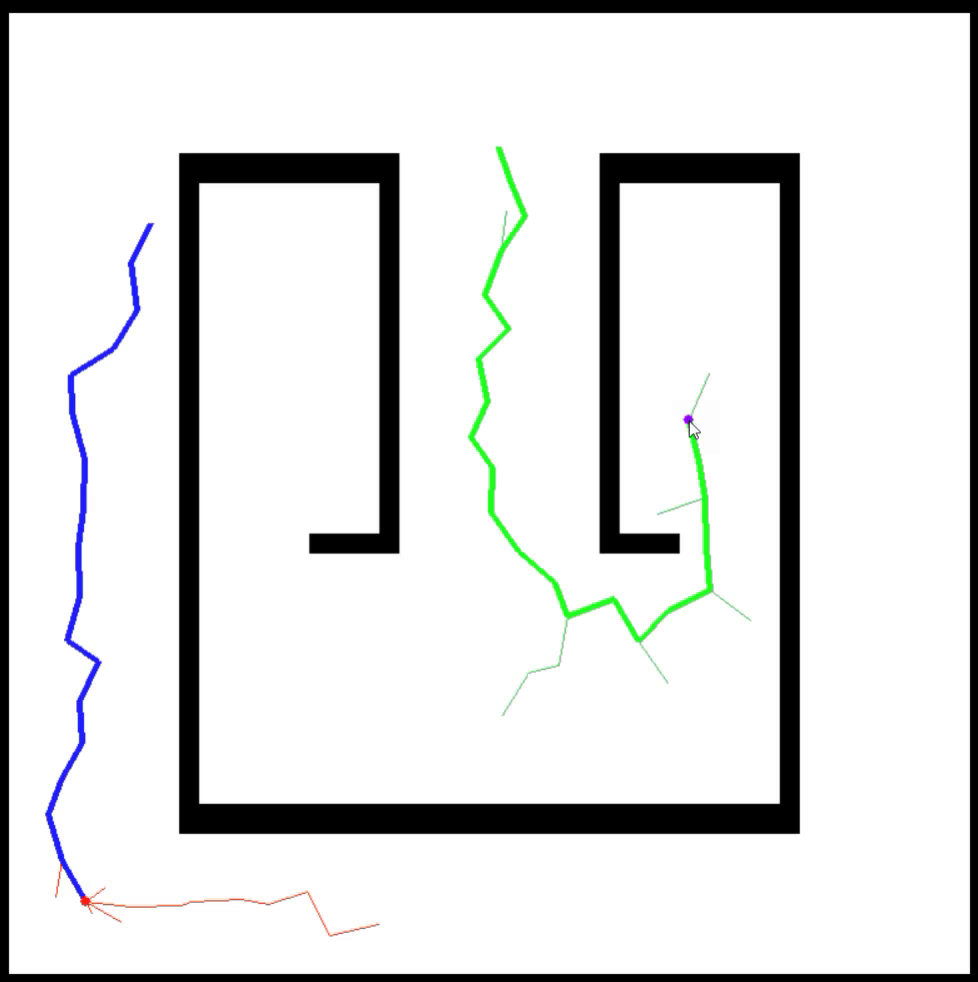}
	}
	\subfigure[]{
		\includegraphics[width=1.6in]{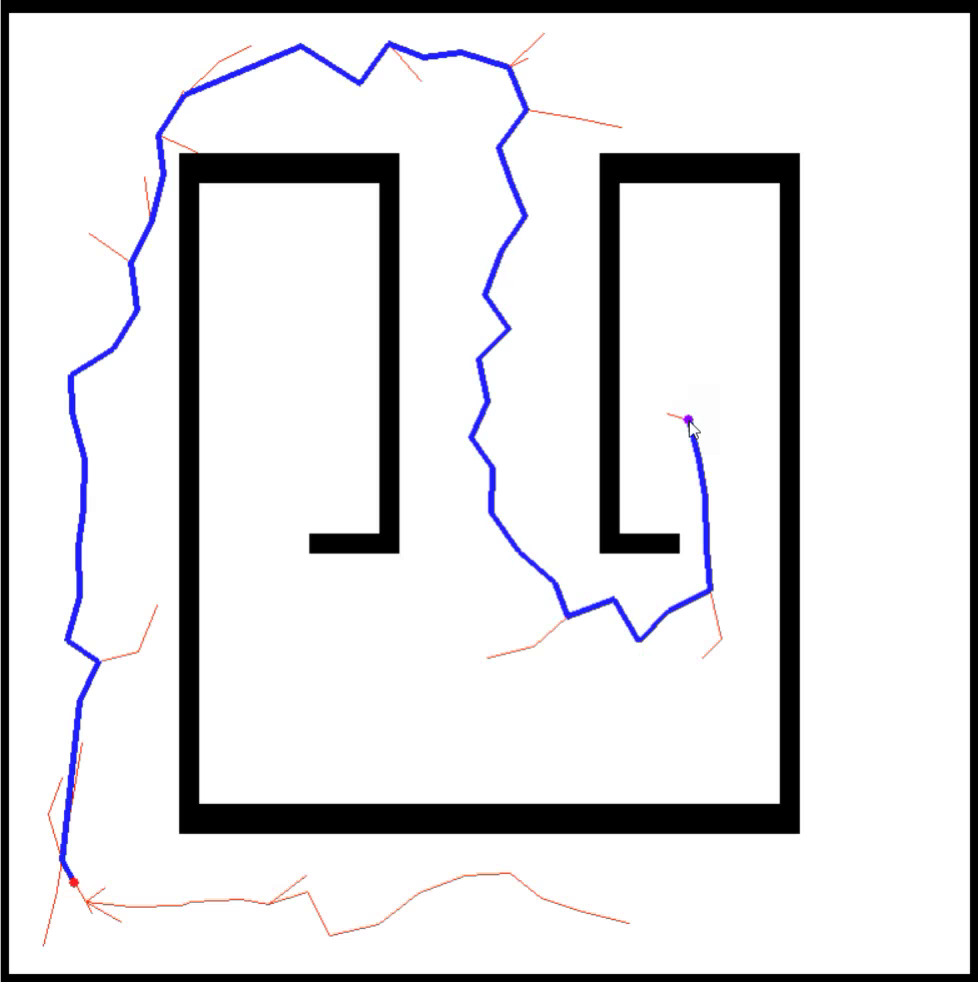}
	}	
	\caption{Bidirectional tree growth rewiring process. (a) The forward tree (in blue) and the reverse tree (in green) grow at the same time. (b) When the two trees are close enough to connect into one tree, the reverse tree stops growing and initializes.}\label{fig2}
\end{figure}

For RRT-based motion planning, the use of a bidirectional search strategy is faster to plan feasible paths than unidirectional.
Fig. \ref{fig2} illustrates the bidirectional tree growth rewiring process. First, the two trees grow simultaneously. When the forward tree and the reverse tree are meet, the forward tree uses the reverse tree to generate the path to the goal while the reverse tree stops growing and initializes. Finally, the forward tree continues to grow to the full map. In this process, the branch information of the forward tree is used for obstacle avoidance and path optimization (refer to Fig. \ref{fig1}). 

Algorithm 1 describes the detail of Bi-AM-RRT*. First, the forward tree and reverse tree information are initialized, and the map information is loaded (Lines 1$\sim$2). Then, the goal points are set and the \textit{X$_{free}$} and \textit{X$_{obs}$} information is continuously updated. The root state of the forward tree follows the position state of the agent, and the goal state is provided by someone. The root state of the reverse tree is set to the goal state, while the goal state is set to the initial state of the agent position and does not change as the agent position moves (Lines 3$\sim$6). When two trees are not connected (i.e., \textit{x$_{goal\underline{~}f}$ $\in$ T$_{r}$, x$_{goal\underline{~}f}$ $\notin$ T$_{f}$}), they grow simultaneously. When two trees are connected successfully, only the forward tree continues to expand to the full map, generating more nodes to optimize the path to avoid obstacles or make the path shorter (Lines 8$\sim$11). The function \emph{Meet}(\textit{T$_{f}$},\textit{T$_{r}$}) denotes that true is returned when the Euclidean distance between \textit{T$_{f}$} and \textit{T$_{r}$} is less than the connection distance $\sigma$ and there is no obstacle blocking it. If true is returned, the information of the reverse tree is fused to the forward tree by the function \emph{Swap}(\textit{T$_{f}$},\textit{T$_{r}$}). Subsequently, the reverse tree stops expanding and initializes (Lines 12$\sim$14). And a collision-free path to the goal is generated, and finally, the agent moves along that path (Lines 15$\sim$16).
When the agent reaches the goal, it waits for information about the next goal. The above steps are repeated once a new goal is given.

\begin{algorithm}[!t]
	\caption{Bi-AM-RRT*}
	\KwIn{\textit{Agent}, \textit{Goal\underline{~}f}, \textit{Map}, \textit{Q\underline{~}f$_{root}$}$\leftarrow$[], \textit{Q\underline{~}r$_{root}$}$\leftarrow$[], \textit{Q\underline{~}f$_{goal}$}$\leftarrow$[], \textit{S\underline{~}f$_{goal}$}$\leftarrow$[]}
	\KwOut{\textit{Path}}
	\textit{Path}$\leftarrow\phi$; \textit{T$_f$}$\leftarrow\phi$; \textit{T$_r$}$\leftarrow\phi$;\\
	\textit{\textit{load()}$\leftarrow$\textit{Map}};\\
	\While{Agent $\notin$ \textit{x$_{goal\underline{~}f}$}}
	{
		\textit{load()$\leftarrow$X$_{free}$, X$_{obs}$};\\
		\textit{x$_{agent}$}$\leftarrow$\textit{Agent},\textit{x$_{root\underline{~}f}$}$\leftarrow$\textit{Agent},\textit{x$_{goal\underline{~}f}$}$\leftarrow$\textit{Goal};\\
		\textit{x$_{root\underline{~}r}$}$\leftarrow$\textit{Goal}, \textit{x$_{goal\underline{~}r}$}$\leftarrow$\textit{x$_{root\underline{~}f}$}(time=0);\\
		\textit{start}$\leftarrow$\textit{clock()};\\
		\While{\textit{clock() - start<t$_{exp}$ $\vee$ x$_{goal\underline{~}f}$}$\neq\phi$}
		{
			\textit{T$_{f}$}$\leftarrow$\textit{Expend\underline{~}f(T$_{f}$, Q\underline{~}f$_{root}$, Q\underline{~}f$_{goal}$, S\underline{~}f$_{goal}$, x$_{goal\underline{~}f}$)};\\
			\If{\textit{x$_{goal\underline{~}f}$ $\notin$ T$_{f}$}}
			{
				\textit{T$_{r}$}$\leftarrow$\textit{Expend\underline{~}r(T$_{r}$, Q\underline{~}r$_{root}$,  x$_{root\underline{~}r}$)};\\
			}
			
		}
		\If{Meet(\textit{T$_{f}$},\textit{T$_{r}$})}
		{
			\textit{Swap(T$_{f}$,T$_{r}$)};\\
			\textit{T$_{r}$}$\leftarrow$\textit{init()};\\
		}
		\textit{Path=Path\underline{~}f(T$_{f}$, Nearest(T$_{f}$, x$_{goal\underline{~}f}$))};\\
		Move Agent towards \textit{x$_{goal\underline{~}f}$};
	}
\end{algorithm}

The tree is grown in a way that maintains the probabilistic completeness of random sampling while using AM for guidance, which make the tree growth more aggressive and efficient. The growth of the forward tree is presented in Algorithm 2. When the goal point is given (Line 1), the forward tree actively grows toward the goal point under the guidance of  the AM. The entire space is then covered by continuous sampling process using the function \emph{SampleState}(T$_{f}$, x$_{goal\underline{~}f}$) (Line 2). \emph{SampleState}(T$_{f}$, x$_{goal\underline{~}f}$) returns the sampling set \textit{X$_{s}$}, which is defined as
\begin{equation}\label{3}
	\begin{split}
		X_{s}=\left\{
		\begin{array}{ll}
			\{x_{goal}\}                    &p>0.7{~}\rm{and}{~}\it x_{goal} \notin T\\
			X_{random} \in X_{free}     &p<0.5{~}\rm or{~}\it x_{goal} \notin T\\
			RewireEllipse(T, x_{goal}) &\rm otherwise \\
		\end{array}
		\right.
	\end{split}
	\nonumber
\end{equation}
where \textit{p} $\in$ [0,1). Afterwards, the root rewiring (see Algorithm 3) is performed to optimize the path (Line 3). In fact, the root rewiring process is always performed. When the goal is found, the rewiring of the goal point  (see Algorithm 5) is then implemented to further optimize the path (Lines 4-5). In particular, the growth of the reverse tree is basically the same as that of the forward tree. Since the reverse tree does not need to reach its own goal, there is no goal rewiring step.

\begin{algorithm}
	\caption{Expend\underline{~}f}
	\KwIn{\textit{Goal}, \textit{Map}, \textit{Q\underline{~}f$_{root}$},  \textit{Q\underline{~}f$_{goal}$}, \textit{S\underline{~}f$_{goal}$}}
	\KwOut{\textit{T$_{f}$}}
	\textit{x$_{goal\underline{~}f}$}$\leftarrow$\textit{Goal};\\
	\textit{T$_{f}$}$\leftarrow$\textit{SampleState(T$_{f}$, x$_{goal\underline{~}f}$)};\\
	\textit{T$_{f}$}$\leftarrow$\textit{RewireRoot(T$_{f}$, Q\underline{~}f$_{root}$)};\\
	\If{{x$_{goal\underline{~}f}$ $\in$ T$_{f}$}}
	{
		\textit{T$_{f}$}$\leftarrow$\textit{RewireGoal(T$_{f}$, Q\underline{~}f$_{root}$, S\underline{~}f$_{goal}$, x$_{goal\underline{~}f}$)};
	}
\end{algorithm}

\subsection{Path Optimization With  Rewiring Strategy}
\begin{figure}
	\centering
	\includegraphics[width=2.7in]{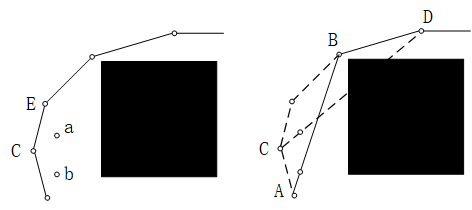}
	\caption{The path optimization process. The tree path is further optimized to A-B on the right when there is a less costly proximity point b around point C. Although the C-D path is better when there is a less costly proximity point a around point E, the path is not optimized due to the obstacle (black square) blocking it.}\label{fig3}
\end{figure}

To optimize the path, a new rewiring method based on RRT* is proposed, which re-searches the grandfather node instead of the parent node to speed up the convergence rate. 
Algorithm 3 provides the root rewiring process of the forward tree. \textit{Q\underline{~}f$_{root}$} is a reference queue used to find less costly points to update \textit{T$_{f}$}. And the new \textit{x$_{root\underline{~}f}$} is the first data of \textit{Q\underline{~}f$_{root}$}. When the root queue is empty, the information of the offset root is added and the root queue is reset (Lines 1$\sim$2). Then, \textit{t$_{root}$} is used to limit the time of root rewiring (Lines 3$\sim$4). When the number of data in \textit{Q\underline{~}f$_{root}$} is greater than 0 and less than or equal to 2, the root rewiring of AM-RRT* \cite{armstrong2021rrt} is used (Lines 5$\sim$6). When the number of data in \textit{Q\underline{~}f$_{root}$} is greater than 2, the proposed new rewiring method is executed for optimization (Lines 7$\sim$8). In such optimization process, the path length is reduced by re-searching for a point near the grandfather node that is less costly and has no obstacle between it and the child node as the new parent node, as shown Fig. \ref{fig3}, which is given in Algorithm 4. The combination of these two rewiring methods accelerates the convergence speed of path optimization and avoids the generation of suboptimal paths at the corners. Thus, the path length is shortened. The root rewiring process in the reverse tree expansion process is the same as that in the forward tree.

\begin{algorithm}
	\caption{RewireRoot}
	\KwIn{\textit{Q\underline{~}f$_{root}$}}
	\KwOut{\textit{T$_{f}$}}
	\If{Q\underline{~}f$_{root}$ = $\phi$}
	{
		\textit{Enqueue(Q\underline{~}f$_{root}$, x$_{root\underline{~}f}$)};\\
		\textit{start}$\leftarrow$\textit{clock()};\\
		\While{clock() - start<t$_{root}$}
		{
			\If{\textit{0<len(Q\underline{~}f$_{root}$)<=2}}
			{
				\textit{RewireRootFirst(T$_{f}$, Q\underline{~}f$_{root}$)};
			}
			\If{\textit{len(Q\underline{~}f$_{root}$)>2}}
			{
				\textit{RewireRootSecond(T$_{f}$, Q\underline{~}f$_{root}$)};
			}
		}
	}
\end{algorithm}

Algorithm 4 summarizes the optimization process when \textit{len(Q\underline{~}f$_{root}$)>2}. In this case, rewiring uses not just the information of \textit{x$_{r1}$} but both \textit{x$_{r1}$} and \textit{x$_{r2}$} to speeds up the path optimization process and reduces the path length. \textit{x$_{r1}$} and \textit{x$_{r2}$} are dequeued in sequence and try to find the point in \textit{x$_{r1}$} nearest neighbor that can reduce the path cost. If it exists, the \textit{T$_{f}$} is updated (Lines 1$\sim$8). If \textit{x$_{near}$} is not in reference queue \textit{Q\underline{~}f$_{root}$}, it is added to the \textit{Q\underline{~}f$_{root}$} (Lines 9$\sim$10).

\begin{algorithm}
	\caption{RewireRootSecond}
	\KwIn{\textit{Q\underline{~}f$_{root}$}}
	\KwOut{\textit{T$_{f}$}}
	\textit{x$_{r1}$}$\leftarrow$\textit{Dequeue(Q\underline{~}f$_{root}$)};\\
	\textit{x$_{r2}$}$\leftarrow$\textit{Dequeue(Q\underline{~}f$_{root}$)};\\
	\For{\textit{x$_{near}$ $\in$ Nearby(T$_{f}$, x$_{r2}$})}
	{
		\If{\textit{FreePath(x$_{near}$, x$_{r1}$)}}
		{
			\textit{c$_{old}$}$\leftarrow$\textit{Cost(T$_{f}$, x$_{r1}$)};\\
			\textit{c$_{new}$}$\leftarrow$\textit{Cost(T$_{f}$, x$_{r1}$)+d$_{E}$(x$_{r2}$, x$_{near}$)+d$_{E}$(x$_{r1}$, x$_{r2}$)};\\
			\If{\textit{c$_{new}$<c$_{old}$}}
			{
				\textit{T$_{f}$}$\leftarrow$\textit{UpdateEdge(T$_{f}$, x$_{near}$, x$_{r1}$)};\\
			}
			\If{\textit{x$_{near}$ $\notin$ Q\underline{~}f$_{root}$}}
			{
				\textit{Enqueue(Q\underline{~}f$_{root}$, x$_{near}$)};\\
			}
		}
	}
\end{algorithm}

When the goal point is in the tree, the goal rewiring method is performed, which is presented in Algorithm 5. In the algorithm, the reference data stack \textit{S\underline{~}f$_{goal}$} and reference queue \textit{Q\underline{~}f$_{goal}$} are used for path optimization, where \textit{S\underline{~}f$_{goal}$} stores the nodes of the current branch and \textit{Q\underline{~}f$_{goal}$} stores the nodes of the next branch. 
When both \textit{S\underline{~}f$_{goal}$} and \textit{Q\underline{~}f$_{goal}$} are empty, the root information of the real-time tree offset is pushed to \textit{S\underline{~}f$_{goal}$} (Lines 1$\sim$2). Here a two-step optimization approach is introduced in this paper: (1) when time is less than \textit{t$_{goal}$} and there is a non-empty set of \textit{len(Q\underline{~}f$_{goal}$)} or \textit{len(S\underline{~}f$_{goal}$)}, the goal rewiring of AM-RRT*\cite{armstrong2021rrt} is performed (Lines 4$\sim$6); and (2) when the time is less than twice \textit{t$_{goal}$} and as long as there is a set \textit{len(Q\underline{~}f$_{goal}$)} or \textit{len(S\underline{~}f$_{goal}$)} longer than 2, the optimization strategy is performed according to Fig. \ref{fig3} (Lines 7$\sim$9), and the details can be found in Algorithm 6.
The two-step optimization approach avoids suboptimal paths caused by obstacles, which reduce path length.

\begin{algorithm}
	\caption{RewireGoal}
	\KwIn{\textit{Q\underline{~}f$_{goal}$, S\underline{~}f$_{goal}$, x$_{goal\underline{~}f}$}}
	\KwOut{\textit{T$_{f}$}}
	\If{Q\underline{~}f$_{goal}$ $\vee$ S\underline{~}f$_{goal}$ = $\phi$}
	{
		\textit{Push(S\underline{~}f$_{goal}$, x$_{root\underline{~}f}$)};\\
		\textit{start}$\leftarrow$\textit{clock()};\\
		\While{clock() - start<t$_{goal}$}
		{
			\If{\textit{len(Q\underline{~}f$_{goal}$)>0 $\wedge$ len(S\underline{~}f$_{goal}$)>0}}
			{
				\textit{RewireGoalFirst(T$_{f}$, Q\underline{~}f$_{goal}$, S\underline{~}f$_{goal}$, x$_{goal\underline{~}f}$)};
			}
		}
		\While{t$_{goal}$<clock() - start<2$\cdot$t$_{goal}$}
		{
			\If{\textit{len(Q\underline{~}f$_{goal}$)>2 or len(S\underline{~}f$_{goal}$)>2}}
			{
				\textit{RewireGoalSecond(T$_{f}$, Q\underline{~}f$_{goal}$, S\underline{~}f$_{goal}$, x$_{goal\underline{~}f}$)};
			}
		}
	}
\end{algorithm}

Algorithm 6 elaborates the second step of the optimization in Algorithm 5. This process uses the information from both points \textit{x$_{r1}$} and \textit{x$_{r2}$} to speed up the optimization process and reduce the path length. When the length of \textit{S\underline{~}f$_{goal}$} 
is greater than 2, \textit{x$_{r1}$} and \textit{x$_{r2}$} are popped in turn, otherwise they exit the queue in turn.
(Lines 1$\sim$ 6). When \textit{x$_{r1}$} is inside the rewire ellipse \cite{gammell2014informed} (nodes inside the ellipse are more likely to be utilized), the cost of each node within the \textit{x$_{r1}$} radius of \textit{e$_{max}$} is calculated. And if there is a point with a smaller cost, the rewiring optimization is performed (Lines 7$\sim$13). If the point is not in \textit{S\underline{~}f$_{goal}$}, it will be added to \textit{S\underline{~}f$_{goal}$} and \textit{Q\underline{~}f$_{goal}$}. Moreover, if the distance from the second node at the top in \textit{S\underline{~}f$_{goal}$} to the goal point is greater than the sum of the distance from \textit{x$_{r1}$} to the goal point and the distance from \textit{x$_{r1}$} to \textit{x$_{r1}$}, the branch is discarded. Then the next iteration is continued (Lines 14$\sim$18).

\begin{algorithm}
	\caption{RewireGoalSecond}
	\KwIn{\textit{Q\underline{~}f$_{goal}$, S\underline{~}f$_{goal}$, x$_{goal\underline{~}f}$}}
	\KwOut{\textit{T$_{f}$}}
	\If{\textit{len(S\underline{~}f$_{goal}$)>2}}
	{
		\textit{x$_{r1}$=Pop(S\underline{~}f$_{goal}$)};\\
		\textit{x$_{r2}$=Pop(S\underline{~}f$_{goal}$)};\\
	}
	\Else
	{
		\textit{x$_{r1}$=Dequeue(Q\underline{~}f$_{goal}$)};\\
		\textit{x$_{r2}$=Dequeue(Q\underline{~}f$_{goal}$)};\\
	}
	\If{\textit{x$_{r1}\in$RewireEllipse(x$_{goal\underline{~}f}$)}}
	{
		\For{\textit{x$_{near}$ $\in$ Nearby(T$_{f}$, x$_{r1}$)}}
		{
			\If{\textit{FreePath(x$_{near}$, x$_{r2}$)}}
			{
				\textit{c$_{old}$}$\leftarrow$\textit{Cost(T$_{f}$, x$_{near}$)};\\
				\textit{c$_{new}$}$\leftarrow$\textit{Cost(T$_{f}$, x$_{r2}$)+d$_{E}$(x$_{r1}$, x$_{near}$) +d$_{E}$(x$_{r1}$, x$_{r2}$)};\\
				\If{\textit{c$_{new}$<c$_{old}$}}
				{
					\textit{T$_{f}$}$\leftarrow$\textit{UpdateEdge(T$_{f}$, x$_{r2}$, x$_{near}$)};\\
				}
				\If{\textit{x$_{near}$ $\notin$ S\underline{~}f$_{goal}$}}
				{
					\textit{S\underline{~}f$_{goal}$}$\leftarrow$\textit{x$_{near}$\}};\\
					\textit{Q\underline{~}f$_{goal}$}$\leftarrow$\textit{x$_{near}$\}};\\
				}
			}
		}
	}
	\If{\textit{len(S\underline{~}f$_{goal}$)>1 $\vee$ d$_{A}$(Second(S\underline{~}f$_{goal}$), x$_{goal\underline{~}f}$) >d$_{A}$(x$_{r1}$, x$_{goal\underline{~}f}$)+d$_{A}$(x$_{r1}$, x$_{r2}$)}}
	{
		\textit{S\underline{~}f$_{goal}$=[]}
	}
\end{algorithm}

The proposed Bi-AM-RRT* can significantly reduce planning costs in both small simple environments and large complex environments with dynamic obstacles. The use of bidirectional tree shorten the time cost of finding the feasible path. During exploration, the suboptimal paths resulting from bidirectional tree connection are optimized by growing the entire path radially around. In addition, the use of the proposed root rewiring and goal rewiring methods accelerates path optimization and reduces the path length.

\section{Experiments and Results}\label{section5}
In order to prove the effectiveness and efficiency of the proposed method, extensive comparative experiments are carried out in different simulation environments. This section gives the
experimental details, while the comparison and discussion of
experimental results are provided.

\subsection{Experimental setting}
The experiments are conducted in PyCharm 2021 on top of a Lenovo Y7000p laptop running Windows OS Intel i5-8300H CPU at 2.3 GHz having 16 GB of RAM. To demonstrate the validity and efficiency of the proposed method, our method is compared with RT-RRT* \cite{naderi2015rt} and AM-RRT* \cite{armstrong2021rrt}. Further, based on the bidirectional search sampling strategy and new rewiring strategy proposed in this work, extensive comparative experiments are designed using five state-of-the-art planners RT-RRT*, RT-RRT*(D), AM-RRT*(E), AM-RRT*(D) and AM-RRT*(G) \cite{naderi2015rt, armstrong2021rrt} to fully evaluate the performance of the Bi-AM-RRT*. Specifically,
\begin{enumerate}
  \item based on five planners, only the bidirectional search strategy is used to design five types of planners, which are denoted as RT-RRT*-1, RT-RRT*(D)-1, AM-RRT*(E)-1, AM-RRT*(D)-1 and AM-RRT*(G)-1.
  \item based on five planners, only the proposed rewiring strategy is used to design five types of planners, which are denoted as RT-RRT*-2, RT-RRT*(D)-2, AM-RRT*(E)-2, AM-RRT*(D)-2 and AM-RRT*(G)-2.
  \item based on five planners, both the bidirectional search strategy and proposed rewiring strategy are used to design five types of planners, which are denoted as Bi-RT-RRT*, Bi-RT-RRT*(D), Bi-AM-RRT*(E), Bi-AM-RRT*(D) and Bi-AM-RRT*(G).
\end{enumerate}

\begin{figure*}
	\centering
	\subfigure[]{
		\includegraphics[width=1.6in]{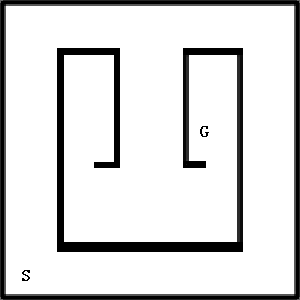}
	}
    \hspace{5mm}
	\subfigure[]{
		\includegraphics[width=1.6in]{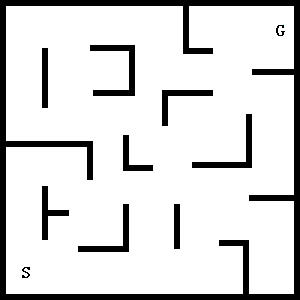}
	}
    \hspace{5mm}
	\subfigure[]{
		\includegraphics[width=1.6in]{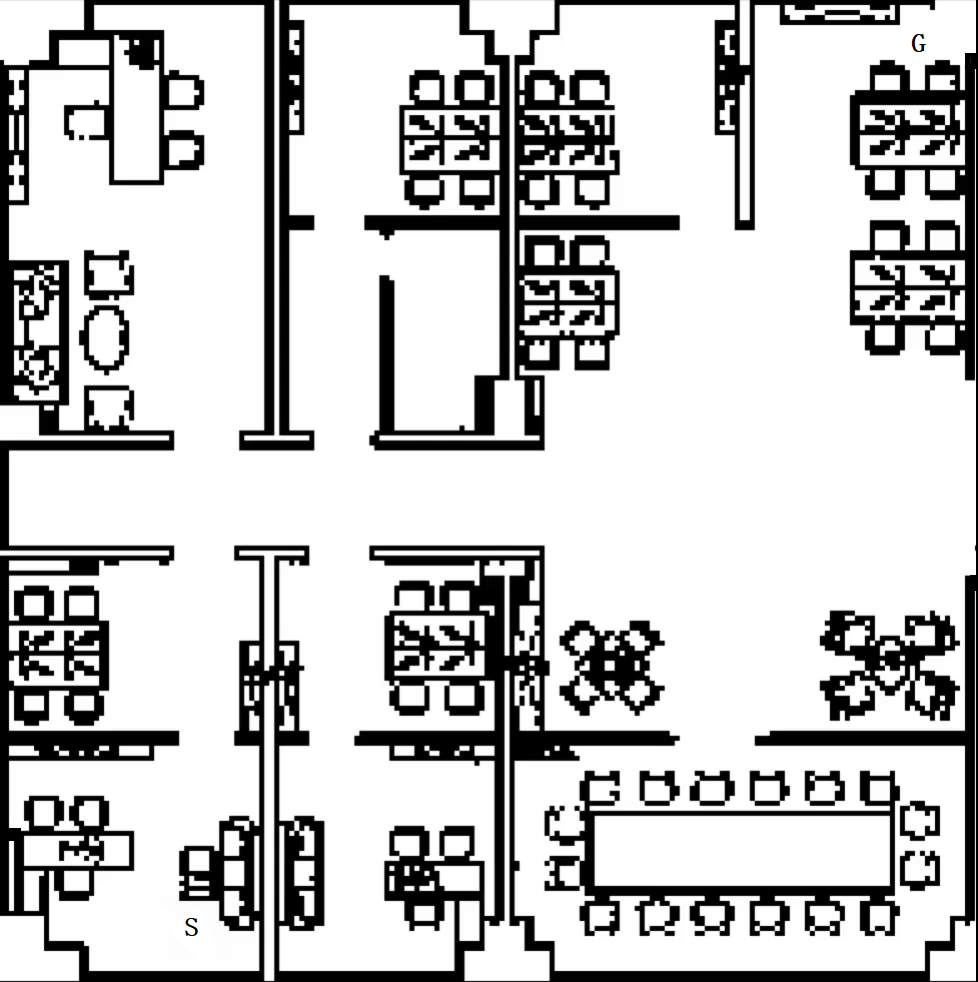}
	}	
	\caption{Experimental scenario: (a) Bug\underline{~}trap, (b) Maze, and (c) Office, where the letters S and G represent the starting point and goal point, respectively. The sizes of the three scenarios are 100$m$ $\times$ 100$m$, 100$m$ $\times$ 100$m$, and 200$m$ $\times$ 200$m$, respectively.}\label{fig5}
\end{figure*}

As shown in Table \ref{table1}, a total of 20 planners are implemented for comparison.
Moreover, experiments are carried out in three challenging scenarios to better demonstrate the robustness and applicability, namely Bug\underline{~}trap, Maze, and Office (see Fig. \ref{fig5}), where the size of Bug\underline{~}trap and Maze is 100$m$ $\times$ 100$m$, and the size of Office is 200$m$ $\times$ 200$m$.
In the three scenarios, the parameter settings of planners are listed in Table \ref{table2}. Note that the connection distance $\sigma$ used in bidirectional tree is set to 50$m$ in the Bug\underline{~}trap scenario and 30$m$ in the other scenarios.

\begin{table*}[!t]
	\renewcommand{\arraystretch}{1.3}
	\centering
	\caption{Description of different planners}\label{table1}
	\scriptsize
	\renewcommand\arraystretch{1.2}
	\begin{tabular}{c c c c c c}
		\toprule
        \multicolumn{1}{c}{\multirow{1}*{Scheme}} & \multicolumn{5}{c}{Planner}  \\
        \midrule
		Original & RT-RRT* & RT-RRT*(D) & AM-RRT*(E) & AM-RRT*(D) & AM-RRT*(G) \\
        Bidirectional search-based & RT-RRT*-1 & RT-RRT*(D)-1 & AM-RRT*(E)-1 & AM-RRT*(D)-1 & AM-RRT*(G)-1 \\		
		Proposed rewiring strategy-based & RT-RRT*-2 & RT-RRT*(D)-2 & AM-RRT*(E)-2 & AM-RRT*(D)-2 & AM-RRT*(G)-2 \\
		Bidirectional-and proposed rewiring strategy-based& Bi-RT-RRT* & Bi-RT-RRT*(D) & Bi-AM-RRT*(E) & Bi-AM-RRT*(D) & Bi-AM-RRT*(G) \\
		\bottomrule
	\end{tabular}
\end{table*}

\begin{table}[!t]
	\renewcommand{\arraystretch}{1.3}
	\centering
	\caption{Parameters setting of planner}\label{table2}
	\scriptsize
	\renewcommand\arraystretch{1.2}
	\begin{tabular}{l c c c c c c}
		\toprule
		& \multicolumn{1}{c}{\textit{t$_{exp}$}/s} & \textit{t$_{root}$}/s & \textit{t$_{goal}$}/s & \textit{e$_{max}$}/m & \textit{n$_{max}$} & \textit{$\sigma$}/m \\
		\midrule
		RT-RRT* & 0.15 & 0.003 & 0.003 & 5 & 12 & 30/50 \\
		RT-RRT*(D) & 0.15 & 0.003 & 0.003 & 5 & 12 & 30/50 \\
		AM-RRT*(E) & 0.15 & 0.002 & 0.004 & 5 & 20 & 30/50 \\
		AM-RRT*(D) & 0.15 & 0.002 & 0.004 & 5 & 20 & 30/50 \\
		AM-RRT*(G) & 0.15 & 0.002 & 0.004 & 5 & 20 & 30/50 \\
		\bottomrule
	\end{tabular}
\end{table}


In the experiment, each planner is tested in a typical task where the agent needs to plan a feasible path to the goal point G from the starting point S in different scenarios with static obstacles, while recording the search time cost and path length of the agent's movement path from the start to the goal. To fairly evaluate the performance of the method, each experiment is repeated 25 times.
The average of the 25 experiments is then used for an unbiased comparison of experimental results. In addition, we further verify the performance of the proposed method in the environment with dynamic obstacles, where dynamic obstacles are simulated by using black circles to block the robot's direction of motion (refer to Fig. \ref{fig1}).

\subsection{Results}

\subsubsection{Scenario With Static Obstacles}
\begin{figure*}
	\centering
	\includegraphics[width=7in]{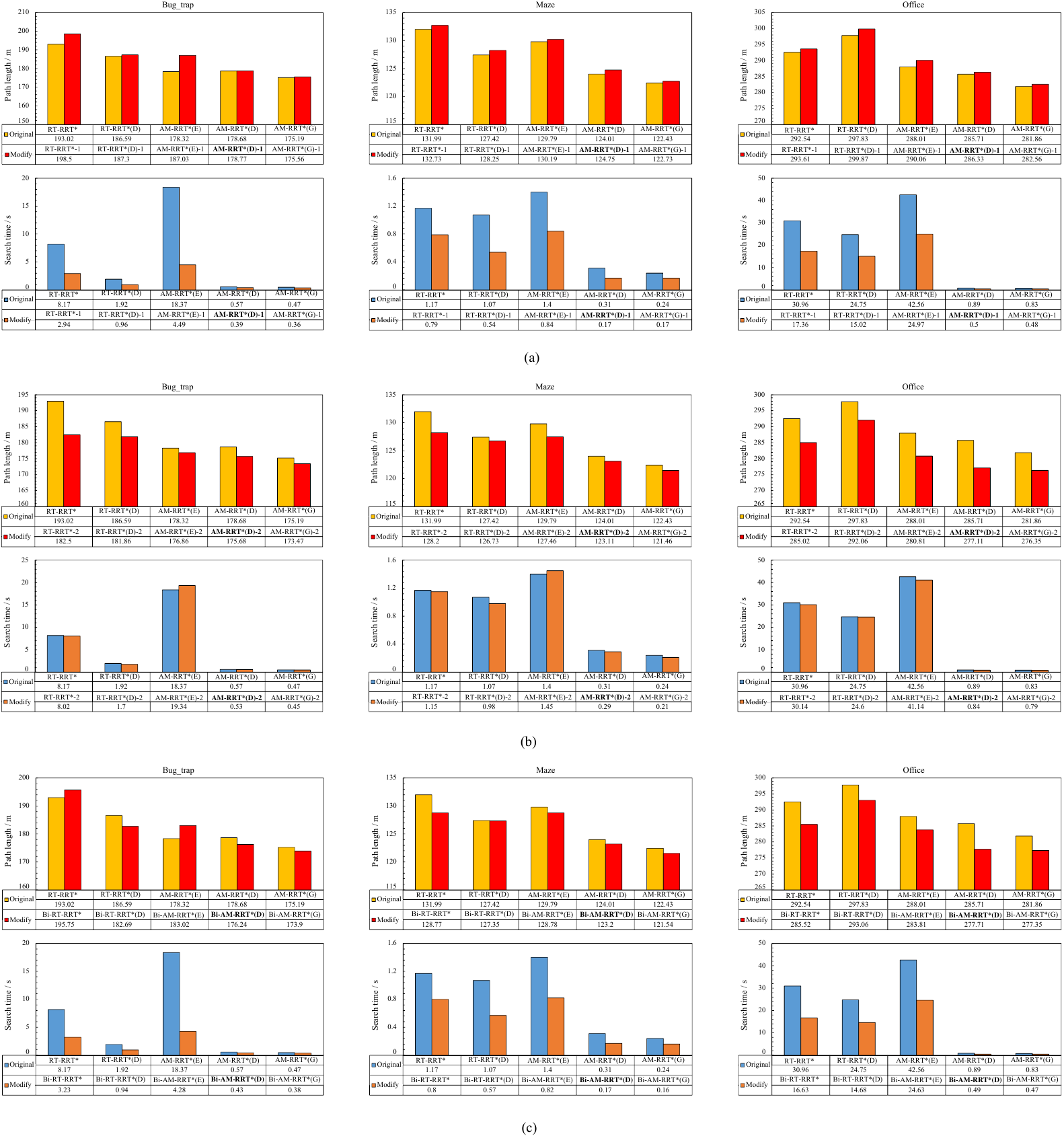}
	\caption{Comparison of experimental results. The average path length and search time required by different planners to find a feasible path from the starting point S to the goal point G in different scenarios, where (a) represents the results with (i.e., Modify) and without (i.e., Original) the bidirectional search sampling strategy, and (b) represents the results with (i.e., Modify) and without (i.e., Original) the proposed rewiring strategy, and (c) represents the results with (i.e., Modify) and without (i.e., Original) the bidirectional search sampling strategy and proposed rewiring strategy.}\label{fig6}
\end{figure*}

\begin{figure}
	\centering
	\subfigure[]{
		\includegraphics[width=1.6in]{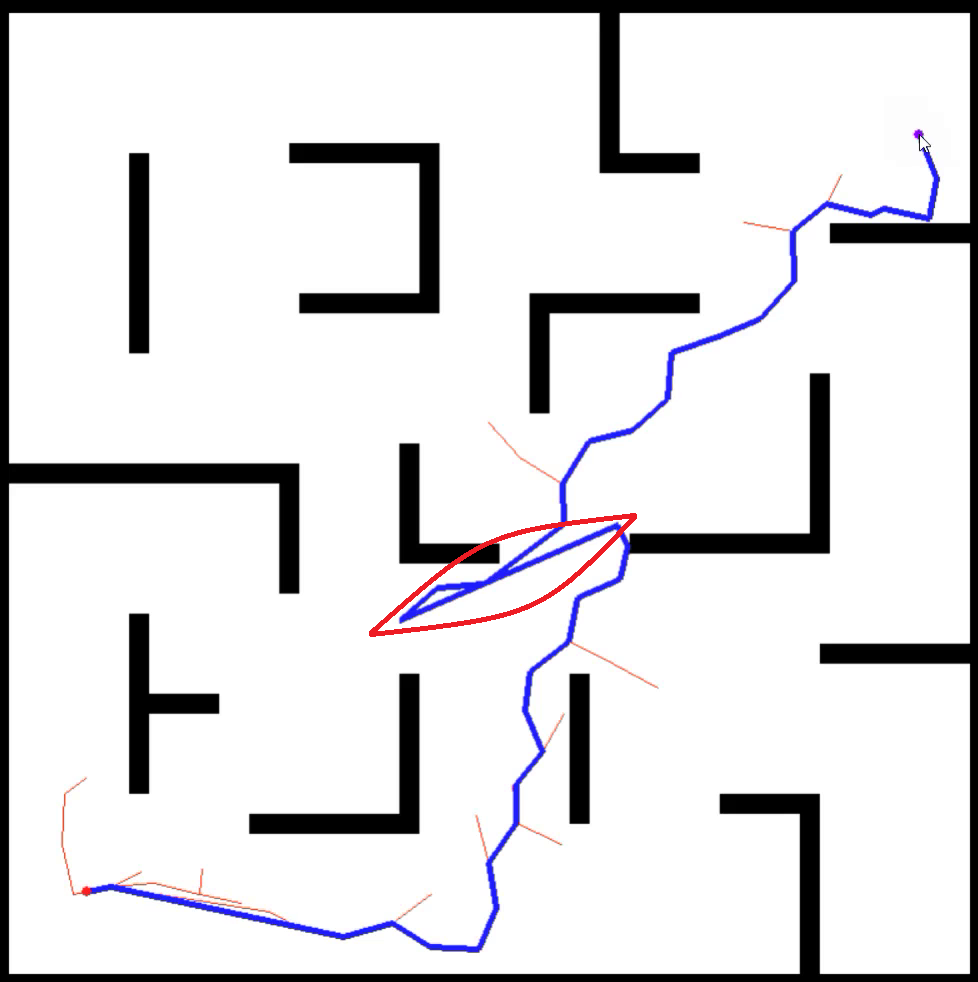}
	}
	\subfigure[]{
		\includegraphics[width=1.6in]{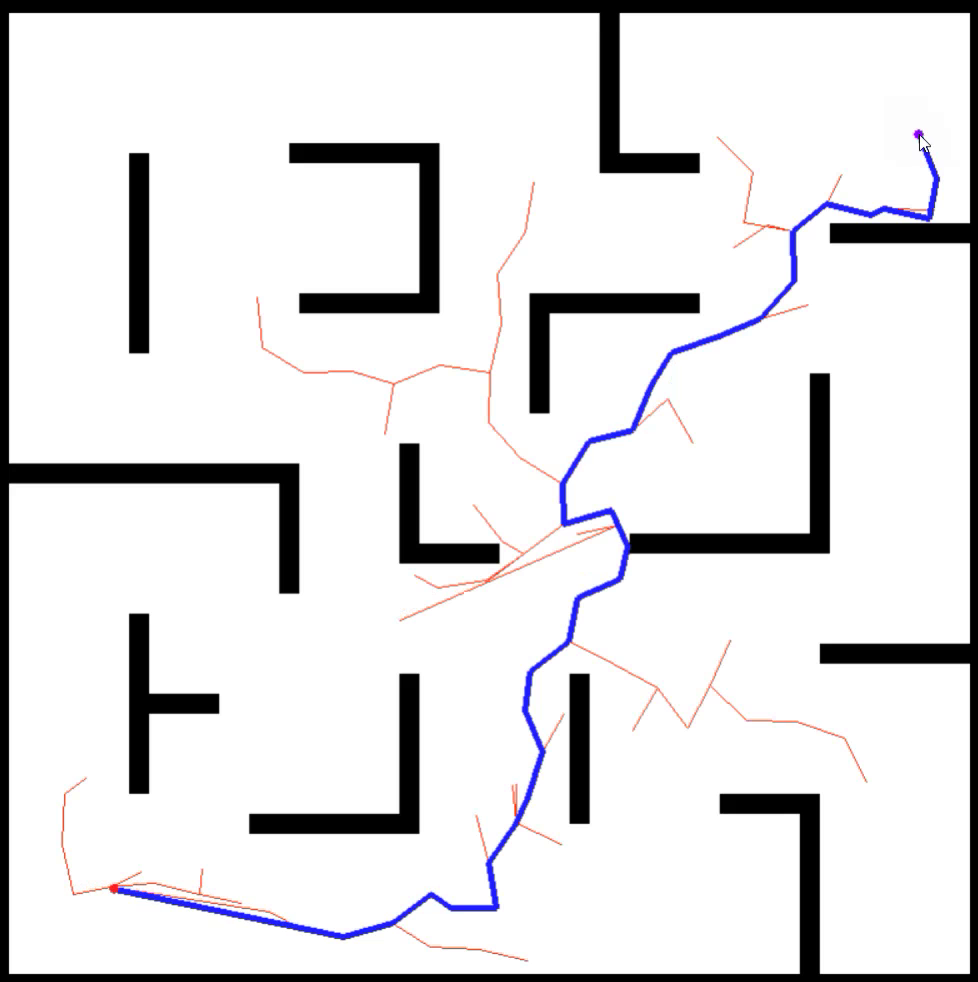}
	}	
	\caption{Path optimization process. (a) When a bidirectional tree connection produces a suboptimal path (red zone), (b) it can be optimized due to the continuous growth of the tree.}\label{fig4}
\end{figure}

According to the experimental setup, 20 different planners were implemented in three different scenarios. The experimental results are shown in Fig. \ref{fig6}. Fig. \ref{fig6}(a) presents the performance comparison comparison with and without the bidirectional search sampling strategy. Since suboptimal paths can be generated in bidirectional tree connections (see Fig. \ref{fig4}), the path length of the five planners based on bidirectional strategy increases, but only by 0.8\%.
Note that the search times are significantly improved with the use of the bidirectional search strategy. In the Bug\underline{~}trap, Maze, and Office, the time costs are reduced by about 69\%, 40.1\%, and 41.7\%, respectively. In particular, the search time of AM-RRT*(E)-1 can be reduced by up to 75.6\% in the Bug\underline{~}trap scenario.
Therefore, the results illustrate that the use of bidirectional search sampling strategy is effective.

The results of Fig. \ref{fig6}(b) demonstrate that the combination of the original method and the proposed rewiring strategy can optimize the average performance of the path length and search time in the three scenarios to a certain extent. And the path length can be reduced by an average of about 2.2\%. Especially for AM-RRT*(D)-2, it can still shorten the path length by 3\% and reduce the search time by 5.6\% even in the large scenario (i.e., Office). Overall, the results demonstrate the effectiveness and generalizability of the proposed rewiring method of this paper.

Fig. \ref{fig6}(c) illustrates a comparison of the results between the solution presented in this paper (i.e., the strategy of fusing bidirectional search sampling strategy and proposed rewiring strategy) and the original solution. It can be seen that the proposed solution can achieve superior performance in terms of path length and search time, except for the slight increase in path length of Bi-RT-RRT* and Bi-AM-RRT*(E) planners in the Bug\underline{~}trap scenario. The reason for this is that the bidirectional strategy and greater connection distance reduce the search time, but when the goal point is found, the number of nodes generated in the tree is insufficient, resulting in a lower degree of path optimization, which will be discussed in detail in Section \ref{section6}.
Besides, on average, Bi-AM-RRT*(E) achieves the superior optimization performance in terms of search time, which can be reduced by 76.7\%, but increased by 2.6\%in terms of path length. It is worth noting that  Bi-AM-RRT*(D) obtains the most promising performance overall. In the three scenarios, Bi-AM-RRT*(D) optimizes search time by 24.6\%, 45.2\% and 44.9\%, respectively, and reduces path length by 1.4\%, 0.7\% and 2.8\%, respectively.

\begin{table}[!t]
	\renewcommand{\arraystretch}{1.3}
	\centering
	\caption{Map processing time for each planner in different scenarios}\label{table3}
	\scriptsize
	\renewcommand\arraystretch{1.2}
	\begin{tabular}{l c c c}
		\toprule
		& \multicolumn{1}{c}{Bug\underline{~}trap} & Maze & Office \\
		\midrule
		Bi-RT-RRT* & / & / & / \\
		Bi-RT-RRT*(D) & 1.5s & 1.5s & 5.6s \\
		Bi-AM-RRT*(E) & / & / & / \\
		Bi-AM-RRT*(D) & 1.5s & 1.5s & 5.6s \\
		Bi-AM-RRT*(G) & 49s & 49s & 232s \\
		\bottomrule
	\end{tabular}
\end{table}

In addition, for Bi-RT-RRT* and Bi-AM-RRT*(E) planners, map processing time is not required. But for Bi-RT-RRT*(D) and Bi-AM-RRT*(D) planners, the diffusion maps are needed, while the geodesic metric is required for Bi-AM-RRT*(G) planner. Table \ref{table3} shows the map processing time for each planner in different scenarios.
Although diffusion map processing takes time, the map processing time for Bug\underline{~}trap and Maze scenarios is about 1.5$s$. Even for larger Office scenario, it only takes 5.6$s$. As tested in this work, the geodesic metric for Bug\underline{~}trap and Maze scenarios take about 49$s$, while Office scenarios take about 232$s$. In this context, 
Bi-AM-RRT*(D) outperforms other planners in terms of total search time (including map processing time) and path length. 
The reason behind this is that diffusion maps are a way to use dimensional collapse to reduce map processing time \cite{coifman2006diffusion, chen2016motion}, such that some details are ignored when processing larger and more complex maps. Therefore, in the Office scenario, the search time of Bi-RT-RRT*(D) is still large, 
but it has less impact on planners based on AM.
Although the comprehensive performance of Bi-RT-RRT*(D) in small scenes is similar to that of Bi-AM-RRT*(D), it is not suitable for larger scenarios. In conclusion, Bi-AM-RRT*(D) is an excellent planner that further improves performance, and is suitable for both small and large scenarios. The results, then, further demonstrate the effectiveness and efficiency of our proposed strategy.

\subsubsection{Scenario With Dynamic Obstacles}
\begin{figure*}
	\centering
	\subfigure[]{
		\includegraphics[width=1.6in]{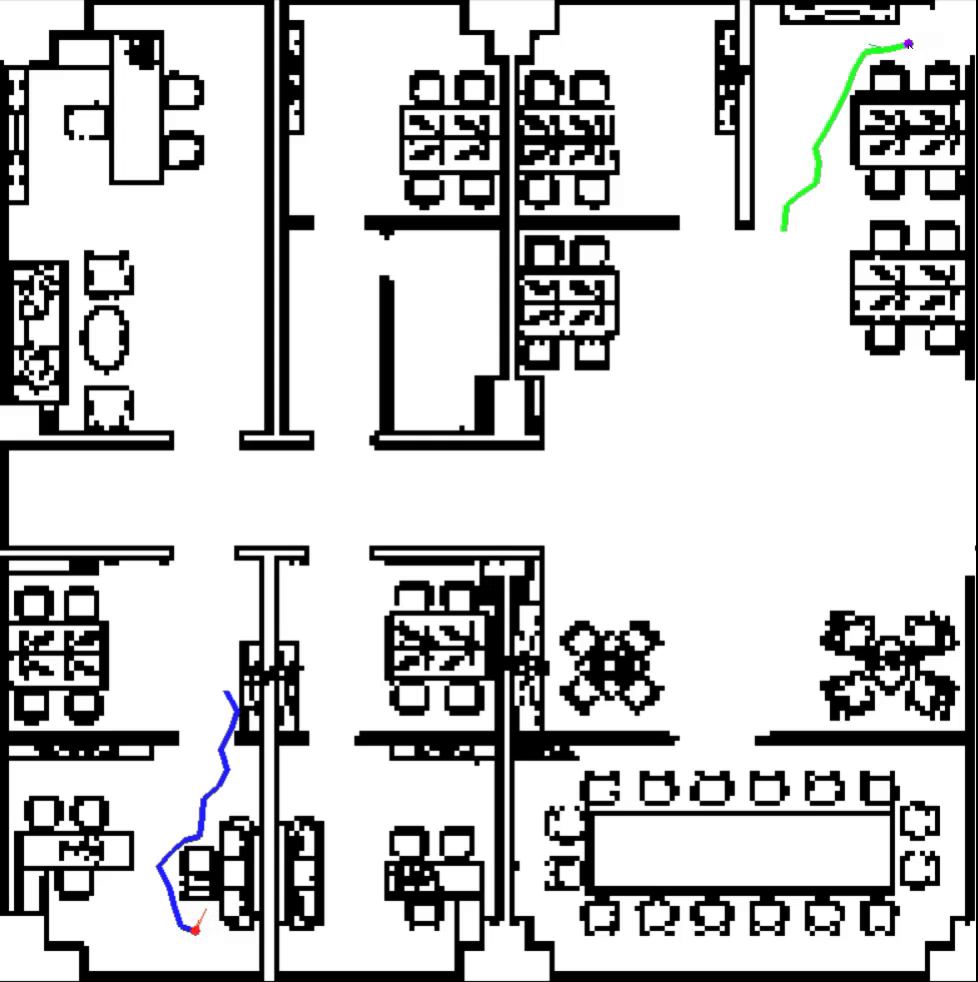}
	}
	\subfigure[]{
		\includegraphics[width=1.6in]{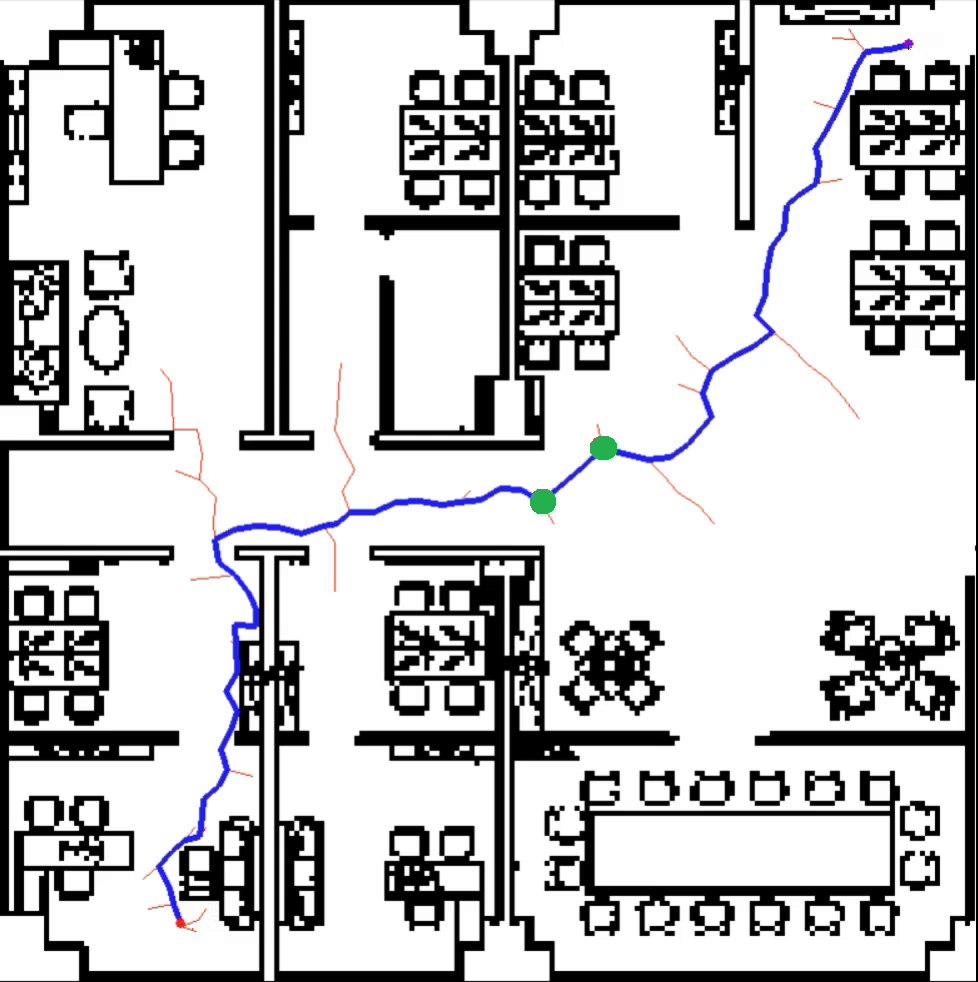}
	}
	\subfigure[]{
		\includegraphics[width=1.6in]{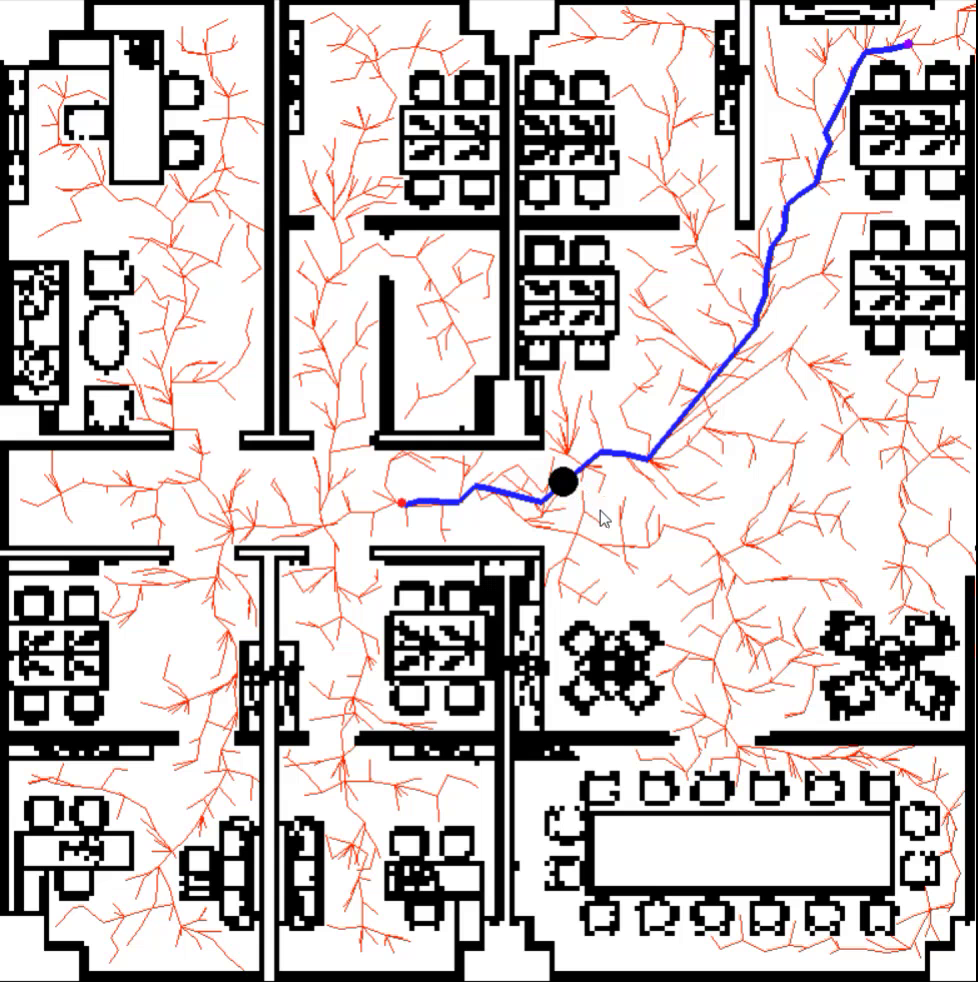}
	}	
	\subfigure[]{
		\includegraphics[width=1.6in]{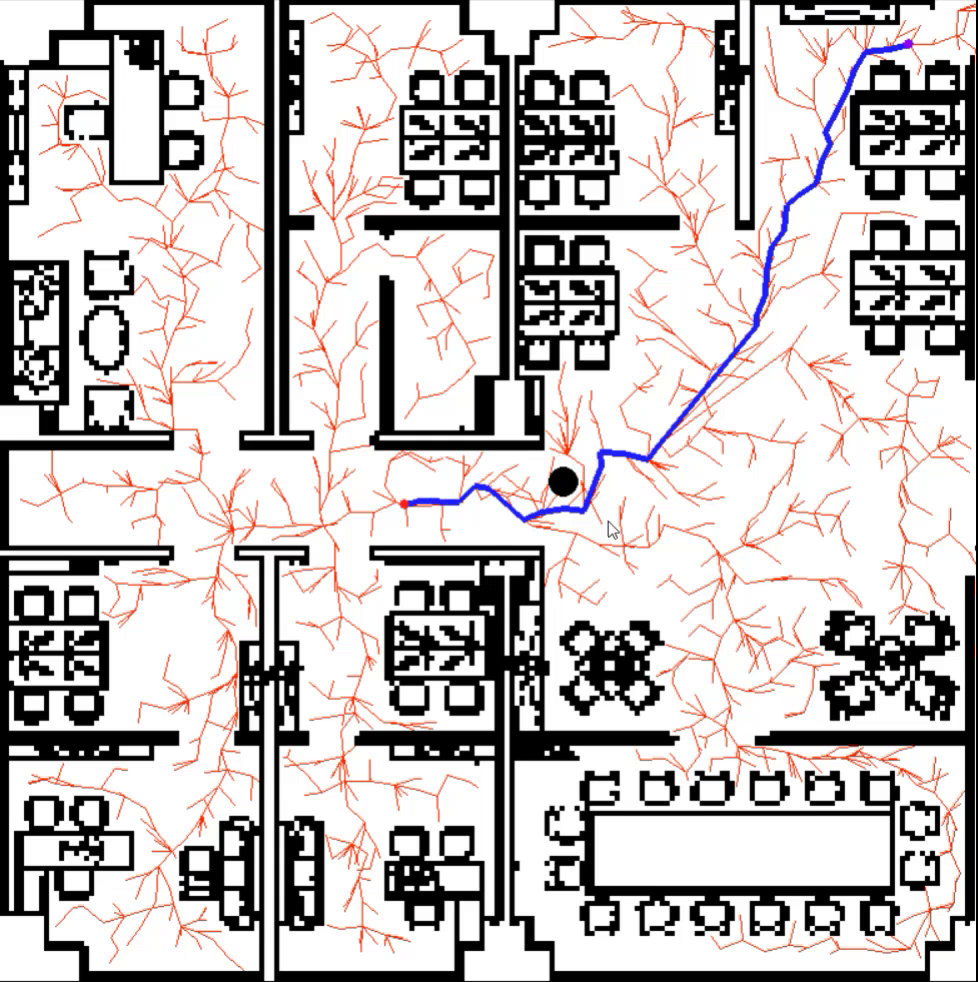}
	}	
	\caption{Obstacle avoidance performance of the proposed algorithm in the Office scenario. In (a) the blue line is the forward tree path and the green line is the reverse tree path. When two trees are close enough, they are connected into one tree through two green points (b). And when the black circle of obstacle appears in the path (c), a feasible path is quickly planned by using the information of nearby branches (d).}\label{fig7}
\end{figure*}

In order to test the obstacle avoidance performance of the proposed method, the experiment is conducted in the Office scenario with the dynamic obstacle. The dynamic obstacle is simulated by using a solid black circle that can be added anywhere at any time to block the path of the robot. The experimental results are depicted in Fig. \ref{fig7}. When the goal point is given, both trees grow at the same time [Fig. \ref{fig7}(a)]. When the distance is close enough [Fig. \ref{fig7}(b)], the two paths are connected to one path at two green dots, and the reverse tree stops growing and initializes. The forward tree uses information from the reverse path to grow quickly to the goal point and to the whole map. During the navigation, when there is an obstacle in the path [refer to Fig. \ref{fig7}(c)], the forward tree uses the node information near the obstacle to quickly generate a feasible path to avoid the obstacle, allowing the agent to move along the planned path and safely reach the goal [see Fig. \ref{fig7}(d)]. Hence the results show that the proposed method can address the obstacle avoidance in dynamic environments.

\section{Disscussion}\label{section6}
This section discusses the effect of the connection distance $\sigma$ on the obstacle avoidance performance and the number of nodes on the path optimization.

\begin{figure}
	\centering
	\subfigure[]{
		\includegraphics[width=1.6in]{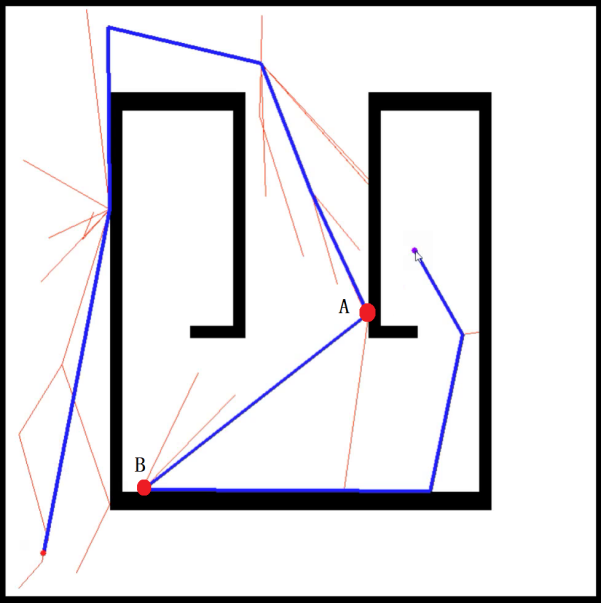}
	}
	\subfigure[]{
		\includegraphics[width=1.6in]{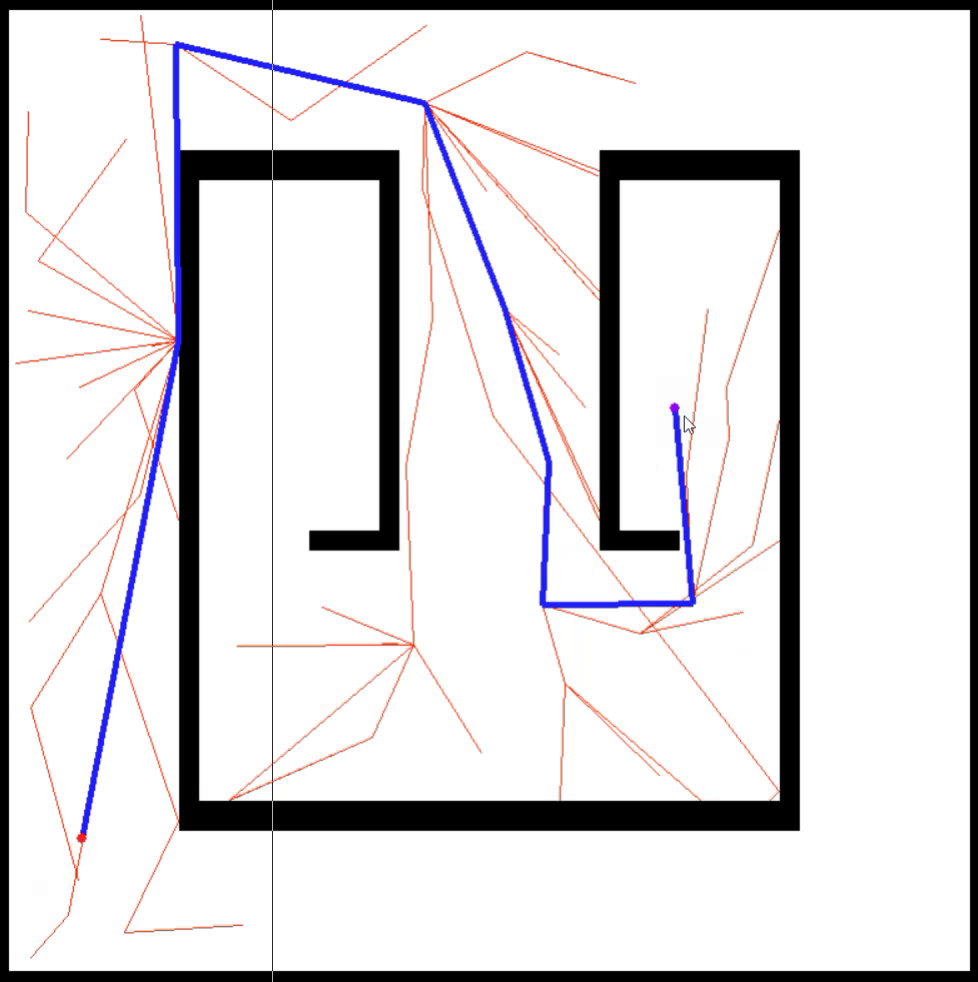}
	}
	\caption{Path planned by Bi-RT-RRT* and Bi-AM-RRT*(E) in the Bug\underline{~}trap scenario. Since the Euclidean metric guides, the forward tree will be trapped in point A, and the reverse tree will be trapped in point B in (a) for a long time, so $\sigma$ is set to 50 for optimization. Although there is a longer suboptimal path after successful connection, but it has been gradually optimized before the agent arrives (b).}\label{fig8}
\end{figure}

In this paper, the values of $\sigma$ are set to 30$m$ and 50$m$. This is very large for the connection distance between the two trees. While this makes it easier to connect the two trees, it also makes it easier to produce the suboptimal path, as shown in Fig. \ref{fig4}. For example, in the Bug\underline{~}trap scenario, $\sigma$ is set to 50$m$ to allow the two trees of the Bi-RT-RRT* and Bi-AM-RRT*(E) planners to connect more easily, as illustrated in Fig. \ref{fig8}. Due to the guidance of Euclidean distance, the two trees are trapped at point A and point B in Fig. \ref{fig8}(a) for a long time, and the distance is far away. This is why the search times of these two planners are optimized by more than 50\%, as shown in Fig. \ref{fig6}(a) and (c). When two trees are connected by the diffusion map or geodesic metric, the other three planners are connected basically in the upper left area of the Bug\underline{~}trap scenario (see Fig. \ref{fig2}). Although $\sigma$ is set at 50$m$, it is not fully utilized, resulting in less optimization of the search time.

\begin{figure}
	\centering
	\subfigure[]{
		\includegraphics[width=1.6in]{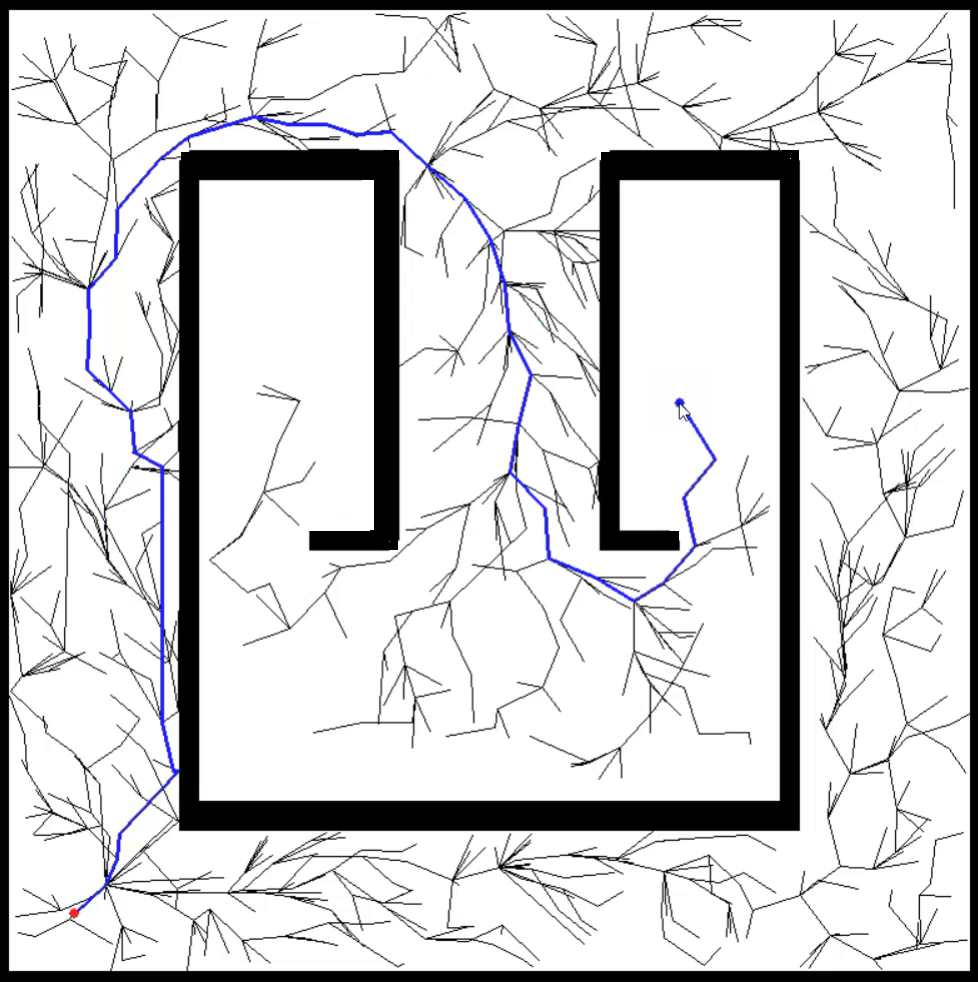}
	}
	\subfigure[]{
		\includegraphics[width=1.6in]{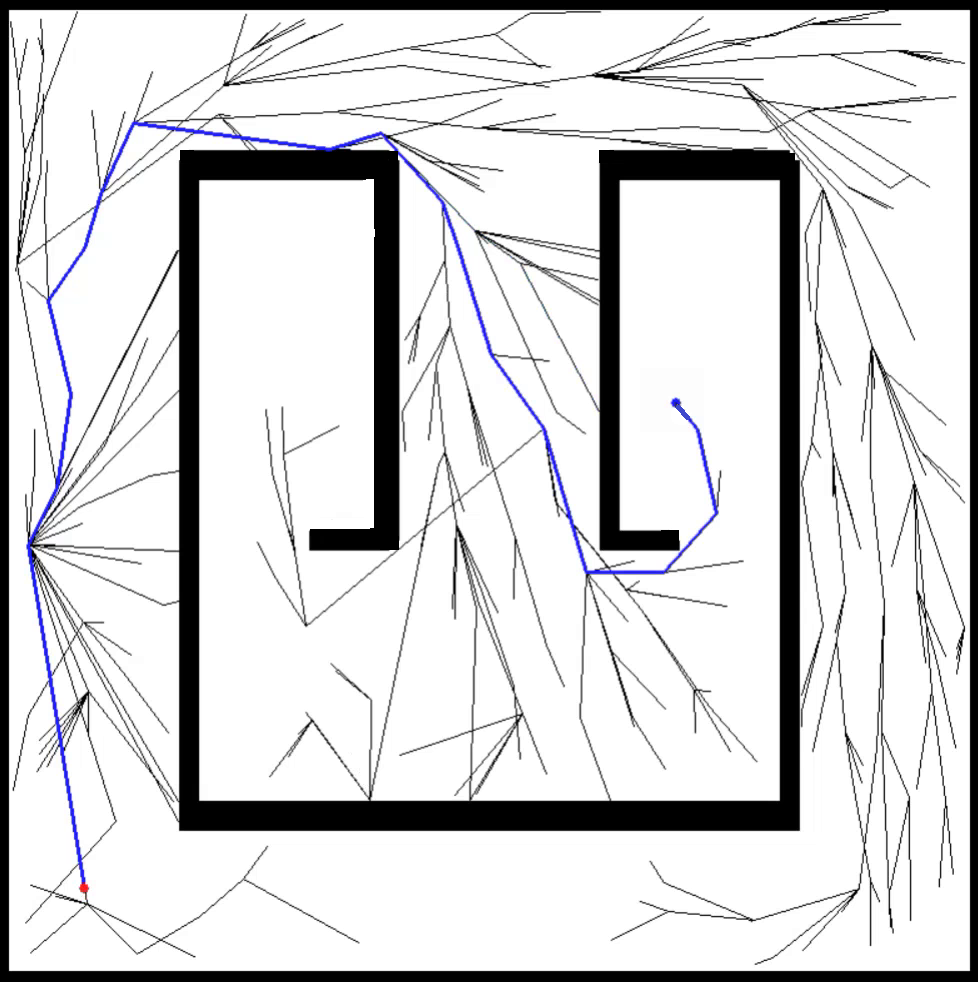}
	}
	\caption{Tree optimization process of the AM-RRT*(E) (a) and RT-RRT* (b).}\label{fig9}
\end{figure}

Also taking Bug\underline{~}trap scenario as an example, it can be seen from Fig. \ref{fig6} that the search time using AM-RRT*(E) is the longest, but the path length is shorter than that of RT-RRT* and almost the same as that of AM-RRT*(D). Since the agent starts moving when the planner finds the goal point, AM-RRT*(E) can generate more nodes in the more search time. In other words, the path can be fully optimized by the time the agent starts moving. Although the tree grows in real time, RT-RRT* does not have enough nodes to path optimization, as depicted in Fig. \ref{fig9}. In the experiment, when a feasible path to the goal point is found, RT-RRT* generates an average of 445 nodes, AM-RRT*(E) generates an average of 1062 nodes, and AM-RRT*(D) generates only 76 nodes on average. After fusing the bidirectional strategy, Bi-RT-RRT* and Bi-AM-RRT*(E) can improve the efficiency of search time by more than 60\%, but lead to a slight increase in path length. Although Bi-AM-RRT*(G) achieves the shortest search time and path length, it requires a long map processing time. Bi-AM-RRT*(D), on the other hand, completes the near-optimal path planning in a less time.

\begin{figure}
	\centering
	\subfigure[]{
		\includegraphics[width=1.6in]{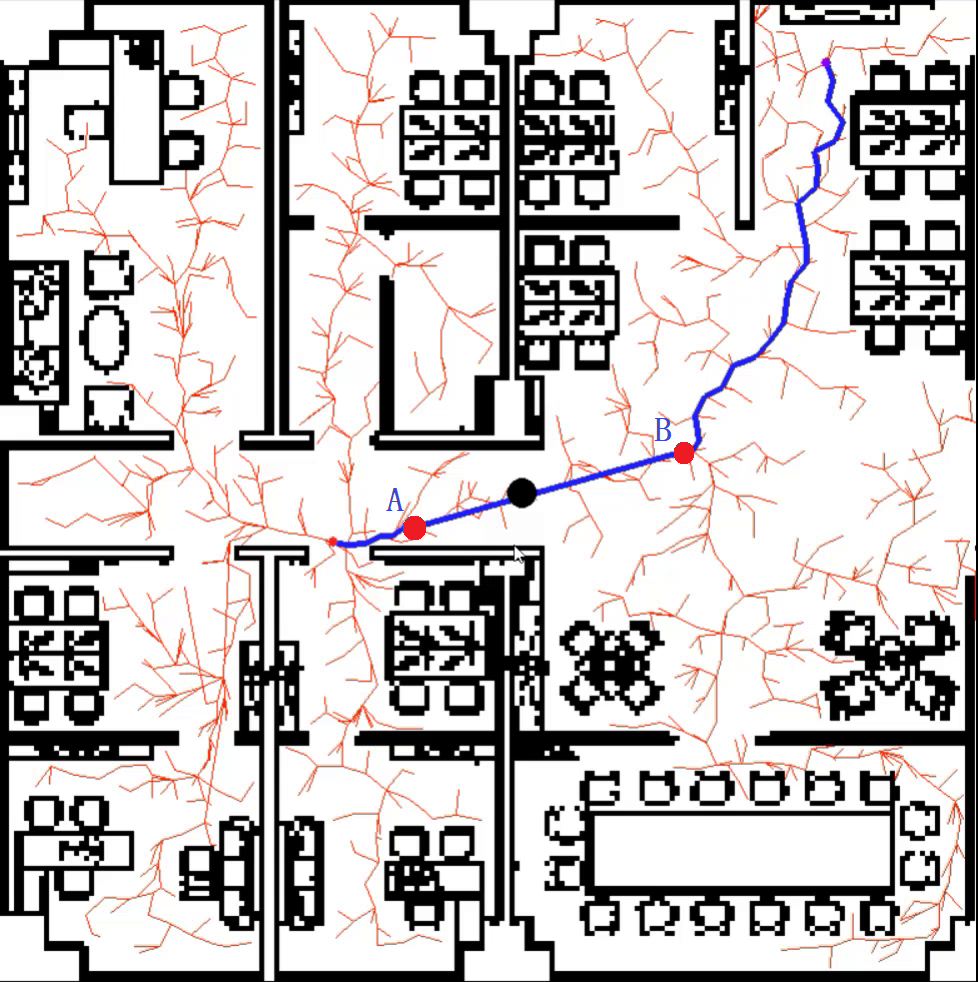}
	}
	\subfigure[]{
		\includegraphics[width=1.6in]{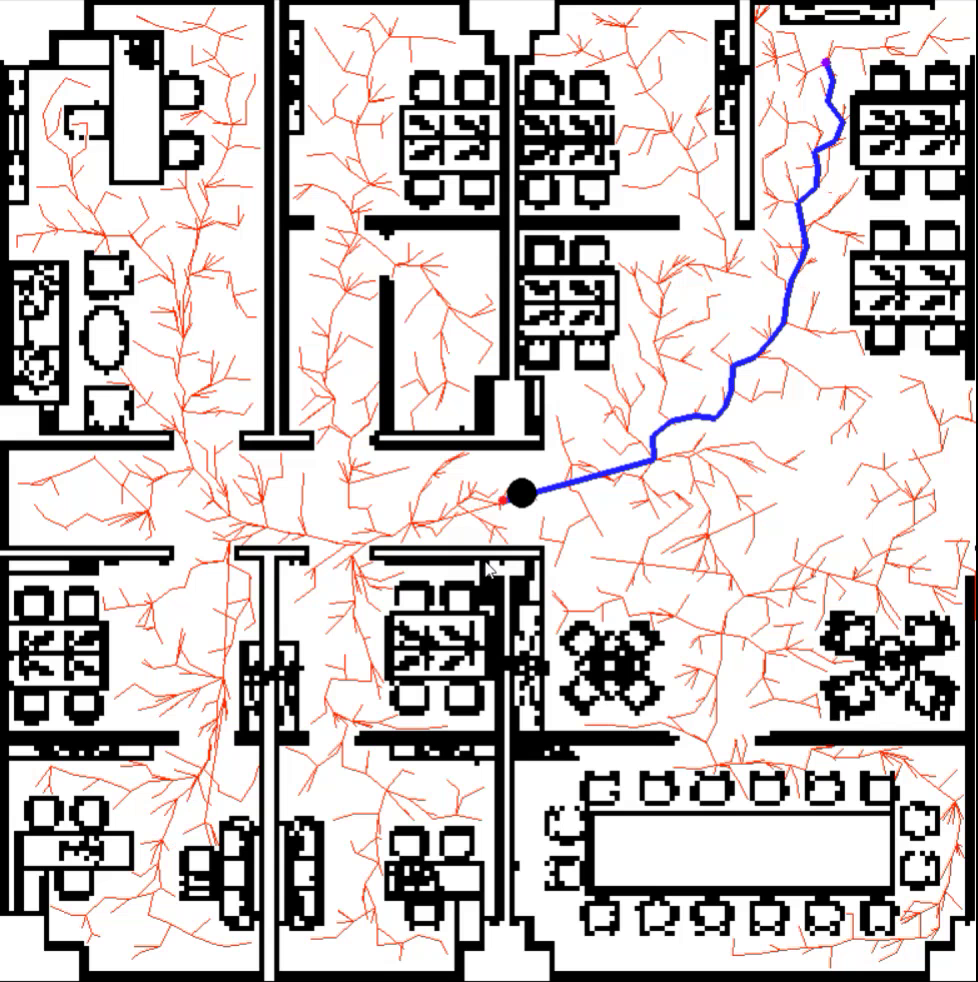}
	}
	\caption{The obstacle avoidance of Bi-AM-RRT*(D) in the Office scenario when $\sigma$ is set to 50$m$. Although the two trees are successfully connected, there are no nodes between the two points A and B for tree growth (a). And the growth rate of the two points A and B, is not enough to maintain the path optimization and obstacle avoidance in that connected path when the agent arrives, resulting in the agent colliding with the obstacle (b).}\label{fig10}
\end{figure}

In the Bug\underline{~}trap scenario, $\sigma$ is set to 50$m$, and all planners can achieve obstacle avoidance performance. Although the three planners do not take full advantage of this large connection range, the Bi-RT-RRT* and Bi-AM-RRT*(E) can maintain path optimization and obstacle avoidance functions. This is due to the loopback path generated after the paths are connected, and there are more nodes available in this region to grow the tree and optimize the structure of the tree by rewiring before the agent arrives. In other scenarios, however, setting $\sigma$ to 50$m$ does not maintain rewiring and obstacle avoidance in some extreme connection situations, as shown in Fig. \ref{fig10}. To this end, the $\sigma$ setting of 30 is tested in the Office scenario, which shows that excellent performance can be maintained even under extreme connection conditions. For this purpose, the $\sigma$ is set to 30$m$ in experiments. The value of $\sigma$ should be determined based on factors such as the size of the scene map, the tree growth time \textit{t$_{exp}$}, and the movement speed of the agent. In this paper, the values of $\sigma$ are not universal in different scenarios, but have certain reference value.

\section{Conclusion} \label{section7}
In this paper, a novel motion planning approach, namely Bi-AM-RRT*, has been proposed. 
Bi-AM-RRT* uses a bidirectional search strategy and a new rewiring approach to reduce the search time and the path length.
In the Bi-AM-RRT*, two trees grow simultaneously when the goal point is not in the forward tree. Then they are connected as one tree when the distance is less than connection distance. In this case, the path to goal is generated by the forward tree while the reverse tree stops growing and initializes. The proposed rewiring method is used to reduce the path length. To this end, the shorter search time allows for faster generation of agent-to-goal paths, which in turn allows for more efficient tree growth by growing trees from points in the path to other regions. Extensive experiments have been carried out in three different scenarios for comparison. The results have demonstrated the validity of our proposal, and effectively improved the motion planning search time and path length. In particular, Bi-AM-RRT*(D) has the best comprehensive performance, while optimizing the search time and path length. In addition, this paper has also discussed the influence of the value of the connection distance on the planner and shown the practicality and robustness of the presented approach.

It is worth noting that in the used bidirectional search strategy, the forward tree only uses the trunk information of the reverse tree, while its branch node information is discarded after a successful connection.
In the future work, the use of branch information will be considered to further improve the path optimization and obstacle avoidance performance. And deploying our solution to mobile robots in real-world scenarios will also be investigated in future research.

\bibliographystyle{IEEEtran}
\bibliography{mybibfile}

\end{document}